%% file: main.tex
\documentclass{article}
\usepackage[preprint]{neurips_2026}

\usepackage{graphicx}
\usepackage{xcolor}

\usepackage{booktabs}
\usepackage{multirow}
\usepackage{tabularx}
\usepackage{makecell}
\usepackage{amsmath,amssymb}

\usepackage{listings}
\usepackage{xcolor}

\lstdefinestyle{prompt}{
  basicstyle=\ttfamily\footnotesize,
  breaklines=true,
  breakindent=0pt,
  columns=fullflexible,
  keepspaces=true,
  numbers=none,
  frame=single,
  framesep=4pt,
  linewidth=\linewidth,
  xleftmargin=0pt,
  xrightmargin=0pt,
  framexleftmargin=4pt,
  framexrightmargin=4pt,
  aboveskip=6pt,
  belowskip=8pt,
  backgroundcolor=\color{gray!8},
  showstringspaces=false,
  literate={_}{\_}1
}

\usepackage{newfloat}
\usepackage{listings}

\usepackage{longtable}
\usepackage{booktabs}
\usepackage{array}

\input{preamble}

\input{metadata}

\begin{document}
\raggedbottom
\maketitle
\input{sections/00_abstract}

\input{sections/01_introduction}

\input{sections/02_related_work}

\input{sections/03_1_ppt_generation_methods}

\input{sections/03_2_data_synthesis}
\input{sections/03_3_sft}

\input{sections/03_4_1_rl_reward}
\input{sections/03_4_2_rl_training}

\input{sections/03_4_3_rl_result}

\input{sections/06_conclusion}
\input{sections/08_AI_use_statement}
\bibliography{references}
\bibliographystyle{plainnat}

\input{sections/07_00_appendix_method_selection}

\input{sections/07_01_appendix_data_synthesis}
\input{sections/07_02_appendix_prompt}

\input{sections/07_03_appendix_sft}
\input{sections/07_04_appendix_sft_ae}

\input{sections/07_05_appendix_rl_reward}
\input{sections/07_06_appendix_rl_training}

\input{sections/07_07_appendix_rl_result}
\input{sections/07_08_appendix_future_work}

\end{document}

%% file: preamble.tex
\usepackage[utf8]{inputenc}
\usepackage[T1]{fontenc}
\usepackage{amsmath,amssymb,amsfonts,mathtools}
\usepackage{booktabs}
\usepackage{graphicx}
\usepackage{float}
\usepackage{algorithm}
\usepackage{algpseudocode}
\usepackage{placeins}
\usepackage{microtype}
\usepackage{nicefrac}
\usepackage{xcolor}
\usepackage{xspace}
\usepackage{url}
\usepackage{hyperref}
\usepackage{multirow}
\usepackage{array}
\usepackage{tabularx}
\usepackage{tikz}
\usetikzlibrary{arrows.meta,fit,positioning,shapes.geometric}

\definecolor{MethodBlue}{HTML}{3B6FB6}
\definecolor{MethodTeal}{HTML}{2A9D8F}
\definecolor{MethodOrange}{HTML}{E69F00}
\definecolor{MethodRed}{HTML}{D55E00}
\definecolor{MethodGray}{HTML}{6B7280}

\hypersetup{
  colorlinks=true,
  linkcolor=MethodBlue,
  citecolor=MethodTeal,
  urlcolor=MethodBlue
}

\newcolumntype{Y}{>{\raggedright\arraybackslash}X}
\newcolumntype{C}[1]{>{\centering\arraybackslash}p{#1}}

%% file: metadata.tex
\title{Building Presentation Intelligence for High-Quality Slide Generation}

\author{%
  \textbf{Yuhang Xie\textsuperscript{1}\quad
  Cheng Li\textsuperscript{2}\quad
  Chenliang Li\textsuperscript{2}\quad
  Mingyu Xu\textsuperscript{2}\quad
  Mingwei Sun\textsuperscript{2}}\\[2pt]
  \textbf{Suhang Zheng\textsuperscript{2}\quad
  Ming Yan\textsuperscript{2}\quad
  Xiaoxiao Xu\textsuperscript{2}\quad
  Lin Qu\textsuperscript{2}}\\[4pt]
  \textsuperscript{1}Peking University\quad
  \textsuperscript{2}Alibaba Group\\[4pt]
  {\small\texttt{yuhangxie@stu.pku.edu.cn}, \texttt{lc497236@alibaba-inc.com}, \texttt{lcl193798@alibaba-inc.com}}\\[2pt]
  {\small\texttt{xumingyu.xmy@alibaba-inc.com}, \texttt{sunmingwei.smw@alibaba-inc.com}, \texttt{suhang.zhengsh@alibaba-inc.com}}\\[2pt]
  {\small\texttt{ym119608@alibaba-inc.com}, \texttt{xiaoxiao.xuxx@alibaba-inc.com}, \texttt{xide.ql@taobao.com}}\\[6pt]
  {\small\raisebox{-0.25em}{\includegraphics[height=1.4em]{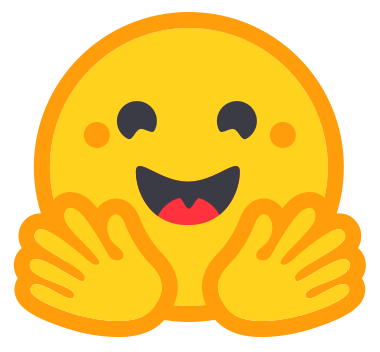}}\enspace\textcolor{orange!75!black}{\textbf{Models}}}\\[2pt]
  {\small
    \href{https://huggingface.co/Logics-MLLM/Logics-PPT-Qwen3.6-35B-A3B-SFT}{\textcolor{orange!75!black}{\underline{Logics-PPT-Qwen3.6-35B-A3B-SFT}}}%
    \enspace\textcolor{orange!75!black}{$\cdot$}\enspace
    \href{https://huggingface.co/Logics-MLLM/Logics-PPT-Qwen3.6-35B-A3B-RL}{\textcolor{orange!75!black}{\underline{Logics-PPT-Qwen3.6-35B-A3B-RL}}}}\\[2pt]
  {\small
    \href{https://huggingface.co/Logics-MLLM/Logics-PPT-Qwen3.6-27B-SFT}{\textcolor{orange!75!black}{\underline{Logics-PPT-Qwen3.6-27B-SFT}}}%
    \enspace\textcolor{orange!75!black}{$\cdot$}\enspace
    \href{https://huggingface.co/Logics-MLLM/Logics-PPT-Qwen3.6-27B-RL}{\textcolor{orange!75!black}{\underline{Logics-PPT-Qwen3.6-27B-RL}}}}
}

%% file: sections/00_abstract.tex
\begin{abstract}
PowerPoint presentations (PPTs) are a dominant medium for communication across diverse domains, yet high-quality PPT generation still depends heavily on labor-intensive human effort. Recent advances in large language models (LLMs) for agentic tasks have created both new opportunities and challenges~\cite{chen2026presentbench} for automatic PPT generation. In this work, we present a unified training framework for automatic PPT generation toward presentation intelligence, integrating cross-validated data synthesis, guided teacher enhancement, and verifiable reward optimization. We first construct a cross-validated pipeline for data synthesis, with explicit control over both difficulty and validity. We then develop a teacher-enhancement framework grounded in rules and prior knowledge to substantially strengthen the instructional capacity of teacher models. Building on this, we further improve supervised fine-tuning (SFT) through a mixed-teacher strategy and a lightweight outcome-driven trajectory filtering method. Finally, based on the resulting SFT models, we introduce a suite of verifiable aesthetic metrics together with a GRPO-based reinforcement learning framework to optimize slide-level visual attributes beyond text-only supervision. Experiments show that our enhanced SFT pipeline improves Qwen-3.6-27B and Qwen-3.6-35B-A3B by 7.45 and 9.35 points on PresentBench, respectively, surpassing GLM-5.2 while maintaining strong generalization. Reinforcement learning further improves performance across five aesthetic dimensions, compensating for the limitations of SFT in visual refinement. Overall, the results demonstrate the effectiveness of our framework for high-quality PPT generation.
\end{abstract}

%% file: sections/01_introduction.tex
\section{Introduction}

\begin{figure*}[t]
    \centering
    \includegraphics[width=0.90\textwidth]{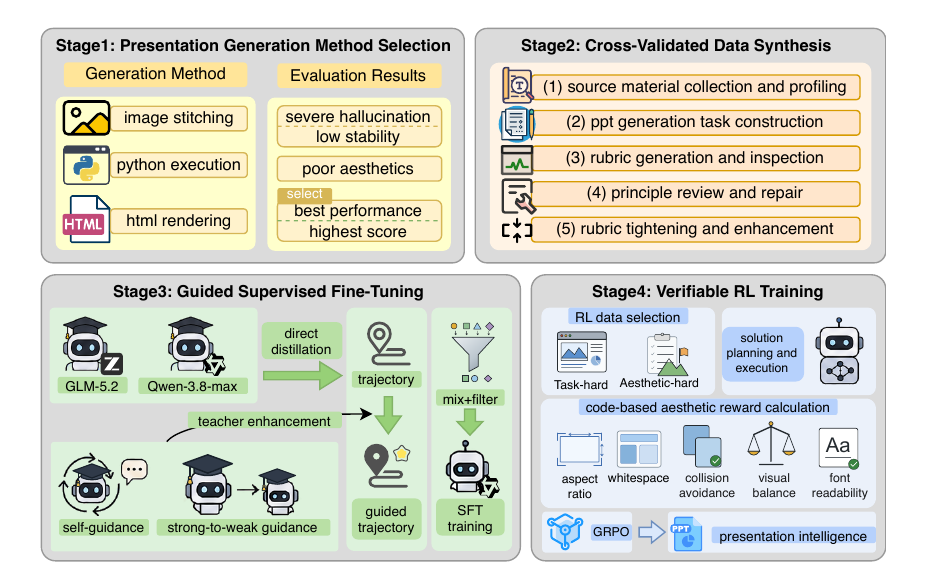}
    \caption{Overview of the proposed presentation generation pipeline.}
    \label{fig:main_fig}
\end{figure*}

% Figure 1: Overview of the proposed presentation generation pipeline. The framework consists of four stages: presentation generation method selection, cross-validated data synthesis, guided supervised fine-tuning with teacher enhancement, and verifiable RL training with code-based aesthetic rewards.

Presentations are a fundamental medium for information delivery across education, business, and research. High-quality presentations require both well-structured content and compelling visual design, enabling audiences to grasp complex ideas efficiently. However, producing such presentations remains labor-intensive and highly skill-dependent, motivating recent efforts to automate the process with large language models (LLMs) and agentic systems. Despite rapid progress in LLM-based agents, slide generation remains a uniquely challenging task because it requires jointly optimizing content quality and visual aesthetics, both of which are inherently subjective and highly domain-sensitive. For instance, academic slides emphasize rigor and precision, whereas advertising slides favor attractiveness and persuasive visual impact. Such heterogeneity makes principled data construction, supervision design, and reward specification particularly difficult.

Existing studies on LLM-based PPT generation have largely focused on agentic framework design, with comparatively limited attention to principled data synthesis, training methodology, and comprehensive verifiable reward design. Methods such as SlideGen~\cite{liang2025slidegen} and PPTAgent~\cite{zheng2025pptagent} introduce multi-agent collaboration and multi-stage planning frameworks tailored to presentation generation, but rely primarily on the intrinsic capabilities of foundation models without task-specific training. MemSlides~\cite{jin2026memslides} and SlideTailor~\cite{zeng2026slidetailor} focus on personalized PPT generation through preference modeling and user profiling, yet likewise do not incorporate an explicit training stage. DeepPresenter~\cite{zheng2026deeppresenter} incorporates a training stage into the generation pipeline, but still relies on a relatively naive SFT paradigm, without strategy adaptation or methodological innovation tailored to the PPT generation setting. SlideRL~\cite{kumar2026learning} incorporates reinforcement learning with specialized tool libraries and LLM-based reward signals, but the resulting supervision remains inherently unverifiable. AeSlides~\cite{pan2026aeslides} studies verifiable aesthetic reinforcement learning for next-slide generation, but does not address the more challenging setting of end-to-end full-presentation generation. In addition, most existing approaches neither provide a systematic account of their data synthesis or collection procedures nor explicitly examine the difficulty and quality of the resulting training data.

To address these limitations, we adopt the comprehensive OpenClaw~\cite{openclaw} framework for PPT generation and develop a high-quality data synthesis pipeline together with efficient and interpretable training strategies (see Figure~\ref{fig:main_fig}). During the SFT stage, we develop a teacher-enhancement framework grounded in explicit rules and prior knowledge, which iteratively alleviates the intrinsic limitations of teacher models and integrates higher-quality expert experience, thereby substantially strengthening their instructional capacity. Building on this foundation, we further improve SFT through mixed-teacher supervision with richer transferable knowledge and an outcome-driven, interpretable trajectory filtering strategy. During the reinforcement learning stage, we design verifiable aesthetic reward metrics to compensate for the weak and indirect aesthetic supervision provided by SFT, thereby markedly improving the first-impression quality of generated presentations. Compared with methods based on unverified metrics, our rewards are more accurate and lead to more effective optimization. Leveraging these metrics, we further develop a GRPO-based reinforcement learning method that directly optimizes the overall quality of PPT generation through verifiable reward signals.

Extensive experiments demonstrate that our three-stage training pipeline, comprising high-quality data synthesis, teacher enhancement, and verifiable rewards, substantially improves both the content quality and visual aesthetics of generated slides. In the SFT stage, our method improves Qwen-3.6-27B and Qwen-3.6-35B-A3B by 7.45 and 9.35 points, respectively. Notably, when trained on only 4,000+ samples, Qwen-3.6-27B already surpasses the substantially larger GLM-5.2. Moreover, although our training queries are synthesized in a style aligned with PresentBench, the resulting models still achieve marked gains on the PPT-generation subset of GDPVal, indicating strong generalization. In the RL stage, both models achieve consistent gains across five aesthetic metrics, compensating for the limited sensitivity of SFT to visual supervision. These results validate the effectiveness of our comprehensive training pipeline.

In summary, our contributions are as follows:
\begin{itemize}
\item We develop a cross-validated data synthesis pipeline for slide generation and release the full codebase for the entire pipeline.

\item We propose an efficient teacher-enhancement framework that compensates for the intrinsic limitations of teacher models and incorporates prior knowledge from stronger models to maximize teacher utility.

\item We introduce a reinforcement learning paradigm with verifiable rewards for slide generation, directly optimizing first-impression visual quality and overcoming the limitations of text-only supervision in SFT.

\item Extensive experiments show consistent improvements on target benchmarks and aesthetic metrics while preserving strong generalization, validating the effectiveness of our comprehensive training pipeline.
\end{itemize}

%% file: sections/02_related_work.tex
\section{Related Work}

\subsection{LLM-Based Slide Generation Frameworks}

Early slide generation systems, such as PPSGen~\cite{hu2014ppsgen} and DOC2PPT~\cite{fu2022doc2ppt}, utilize LLMs for content summarization while relying on predefined templates for text filling and visual layout construction. Although these approaches reduce the capability requirements imposed on LLMs, they suffer from limited flexibility and struggle to produce presentations comparable to human-designed slides. With the rapid development of agentic LLM systems, recent works have shifted toward multi-agent collaboration and multi-stage planning frameworks for PPT generation. PPTAgent~\cite{zheng2025pptagent} introduces an editing-based two-stage framework that first plans presentation content and subsequently performs iterative refinement. PreGenie~\cite{xu2025pregenie} adopts collaborative multimodal large language models (MLLMs) to generate an initial draft followed by review-driven regeneration. DeepPresenter~\cite{zheng2026deeppresenter} further incorporates environment-grounded reflection to mitigate defects that cannot be reliably identified through internal reasoning signals alone. These frameworks substantially improve generation flexibility and task decomposition capability, but still primarily depend on the intrinsic reasoning ability of foundation models.

\subsection{Training Strategies for Slide Generation}

Beyond framework design, recent studies have also explored task-specific training strategies for PPT generation. AutoPresent~\cite{ge2025autopresent}, SlideCoder~\cite{tang2025slidecoder}, and DeepPresenter~\cite{zheng2026deeppresenter} improve generation quality through supervised fine-tuning on their respective frameworks. However, existing SFT-based approaches still rely heavily on heuristic data construction and largely empirical training pipelines, without making corresponding adaptations to the unique requirements of PPT generation. Prior work has neither explicitly enhanced the instructional capacity of teacher models for this task nor systematically validated the reliability of distilled data in the PPT generation setting. In reinforcement learning, SlideRL~\cite{kumar2026learning} introduces six reward dimensions, including inverse specification and content-related signals, but does not verify the reliability of its LLM-based reward scores. AeSlides~\cite{pan2026aeslides} further proposes rule-based aesthetic metrics, yet focuses on next-slide generation rather than the more practical setting of full deck generation. Overall, training methodologies for end-to-end PPT generation remain relatively underexplored.

%% file: sections/03_1_ppt_generation_methods.tex
\section{Presentation Generation Method Selection}

Existing LLM-based approaches to automatic presentation generation can be broadly grouped into three paradigms, namely Python-based code generation~\cite{ge2025autopresent,jung2025talk}, HTML-based rendering~\cite{pan2026aeslides,xu2025pregenie}, and direct image generation~\cite{betker2023improving}. These paradigms differ substantially in flexibility, controllability, and extensibility, and thus exhibit distinct strengths and limitations. Through empirical analysis, we find that image-based generation often suffers from severe textual hallucinations, weaker controllability and greater computational cost. By contrast, the rapidly improving code generation capability of LLMs shows much stronger potential for reliable presentation synthesis. We therefore focus on the two code-centric paradigms.

\begin{table}[t]
\centering
\small
\setlength{\tabcolsep}{5pt}
\caption{Results on PresentBench and the PPT subset of GDPVal under four generation settings.}
\label{tab:gdpval_presentbench_avg}
\begin{tabular}{llcccc}
\toprule
\textbf{Dataset} & \textbf{Model} & \textbf{Single-HTML} & \textbf{Multi-HTML} & \textbf{Single-Python} & \textbf{Multi-Python} \\
\midrule
\multirow{2}{*}{PresentBench}
& Qwen-3.6-27B     & \textbf{70.45} & 63.39 & 59.56 & 60.80 \\
& Qwen-3.6-35B-A3B & \textbf{64.93} & 64.28 & 64.10 & 64.91 \\
\midrule
\multirow{2}{*}{GDPVal (PPT)}
& Qwen-3.6-27B     & \textbf{74.49} & 74.12 & 66.77 & 68.31 \\
& Qwen-3.6-35B-A3B & \textbf{74.01} & 71.58 & 70.03 & 71.57 \\
\bottomrule
\end{tabular}
\end{table}

\paragraph{Evaluation Setup.}
We evaluate Qwen-3.6-27B and Qwen-3.6-35B-A3B within the OpenClaw framework. Beyond two code-centric paradigms, we further extend the setting to single-page and multi-page generation, where each file corresponds to a single slide or a full presentation deck, capturing the trade-off between slide-level refinement and global coherence. Experiments are conducted on PresentBench~\cite{chen2026presentbench} and the presentation-generation subset of GDPVal~\cite{patwardhan2026gdpval}. HTML outputs are rendered into rasterized PDFs via browser screenshots, while Python outputs are executed and converted to PDF using headless LibreOffice. All outputs are evaluated under a unified protocol with Gemini-3-Flash-Preview~\cite{google2026gemini3flashpreview}, which scores each PDF independently on every predefined rubric. This per-rubric protocol reduces the evaluation degradation caused by long contexts and cross-rubric interference.

\paragraph{Evaluation Results.}
As shown in Table~\ref{tab:gdpval_presentbench_avg}, the Single-HTML setting consistently delivers the best overall performance for both Qwen-3.6-27B and Qwen-3.6-35B-A3B. Across the two code-centric generation paradigms, we observe that HTML outputs are markedly superior to Python outputs in visual design quality. Specifically, HTML exhibits clear advantages in layout composition, color design, visual diversity, and mathematical rendering. In particular, since python-pptx lacks a native mathematical typesetting engine, formulas can only be rendered as plain text, which substantially limits expressive fidelity in formally structured presentations. Between the single-page and multi-page settings, we further find that single-page generation is more robust and easier to optimize. Independent slide-wise generation naturally facilitates fault isolation, precise error localization, and incremental refinement, whereas multi-page generation must maintain all slides and numerous implementation details within a single file, making it more susceptible to blank pages and construction errors. Based on these findings, we adopt Single-HTML, the most promising paradigm in our evaluation, as the presentation generation setting for subsequent training.

%% file: sections/03_2_data_synthesis.tex
\section{Cross-Validated Data Synthesis}
\label{sec:data-synthesis}

To support automatic PPT generation at scale while explicitly controlling the difficulty of synthesized data, we construct not only tasks but also instance-level rubrics. The main challenge is that naive automatic synthesis often produces low-complexity tasks that fail to reflect realistic usage scenarios, together with rubrics that are structurally plausible yet weak in discrimination. This issue is especially severe when generation and validation are entangled within the same process, allowing artifacts to be judged by the same mechanism that created them. To address this problem, we develop a cross-validated synthesis pipeline in which task construction, rubric construction, and quality control are defined through cross-stage constraints rather than self-certification. This design improves validity, mitigates answer leakage, and makes task difficulty and rubric coverage explicitly controllable. Using this pipeline, we constructed 4,802 instances with 209,966 rubric items, spanning 25 primary and 162 secondary domains.

Concretely, our pipeline treats data synthesis as a process of externally grounded co-construction rather than isolated artifact generation. Each source document is first abstracted into a structured source card that serves as the evidential substrate for all subsequent stages. Tasks are then derived from scope-level information only, while answer-critical evidence is deliberately withheld, so that task validity is defined by non-disclosure with respect to the rubric answer space rather than by instruction specificity alone. Based on the same evidential substrate, instance-level rubrics are constructed to remain complementary to the task while being explicitly grounded in source evidence, with completeness and correctness separated as distinct verification views. The synthesized task--rubric pairs are subsequently audited as coupled objects to enforce cross-stage consistency, structural legality, and evidential coverage, and the final rubric is further tightened without changing its measurement structure, thereby increasing discriminative power while preserving comparability. In this formulation, external validity does not arise from self-consistency within any single stage, but from the mutual constraints linking source, task, and rubric, making answer non-disclosure, evidential grounding, and difficulty control explicit properties of the synthesis process.

%% file: sections/03_3_sft.tex
\section{Guided Supervised Fine-Tuning}

\subsection{Data Distillation Method}

PPT generation is a long-horizon and highly challenging task that entails document understanding, information acquisition, content organization, and visual planning, among other stages. Even strong models often struggle to produce presentations and distilled trajectories that simultaneously satisfy multiple quality criteria. To obtain higher-quality supervisory data, we improve trajectory distillation from two complementary perspectives. First, we propose \textbf{self-guided distillation}, following an iterative loop of \emph{small-scale distillation $\rightarrow$ trajectory and output analysis $\rightarrow$ principle induction $\rightarrow$ guided redistillation}. This process progressively reveals the failure modes of the teacher model itself and distills corresponding remedies into explicit guiding principles for subsequent rounds of distillation. Second, we adopt a \textbf{strong-to-weak guidance} paradigm. By comparing the trajectories and outputs of stronger and weaker teacher models, we identify the strengths of stronger models in planning, grounding, and slide organization, and encode them as explicit guidance for weaker models. In this way, we enhance the supervisory effectiveness of weaker yet substantially more cost-efficient teacher models for student training, thereby reducing the overall training cost.

Specifically, we use Qwen-3.8-Max~\cite{qwen3.8} and GLM-5.2~\cite{zeng2026glm} as teacher models for trajectory distillation. Qwen-3.8-Max is a state-of-the-art proprietary model for general-purpose agentic reasoning and coding, providing a strong upper-bound reference for transferring broad capabilities to student models. GLM-5.2 is specifically optimized for slide generation and exhibits stronger task specialization in presentation structuring, page organization, and instruction following. Their combination provides complementary supervision spanning both general capability and domain expertise. For Qwen-3.8-Max, we apply self-guided distillation by augmenting the original task prompt with explicit constraints on content and design principles. For GLM-5.2, we jointly employ both guidance strategies and construct three levels of guidance:

\begin{itemize}
    \item \textbf{Basic Guidance (Base)}: The prompt contains only the original benchmark task description together with the required PowerPoint generation format.
    \item \textbf{Self-Guidance (Pro)}: Built on Base, this variant adds strict source-grounding requirements for material-supported content, 15 anti-hallucination principles for content fidelity, and 6 layout-related design constraints.
    \item \textbf{Self-Guidance with Strong-Model Augmentation (Max)}: Built on Pro, this variant further incorporates a Consistent Presentation Design System, additional visual quality constraints, and an iterative Render--Inspect--Fix verification loop.
\end{itemize}

This design enables a systematic analysis of how instruction granularity and guidance strength affect the quality of distilled trajectories and, consequently, downstream student performance.

\subsection{Training Method}

\paragraph{Implementation Details.}
We conduct SFT on two backbones, Qwen-3.6-27B (dense)~\cite{qwen3.6-27b} and Qwen-3.6-35B-A3B (MoE)~\cite{qwen3.6-35b-a3b}. All models are fine-tuned in bf16 with the Adam optimizer, using a peak learning rate of 5e-6, cosine decay, weight decay of 0.01, and gradient clipping at 1.0. Training is performed on 64 GPUs with a per-device batch size of 1 and gradient accumulation of 4, resulting in a global batch size of 4. The dense model is trained for 5 epochs, while the MoE model is 

\paragraph{Training Data.} Following the distillation protocol, we construct the SFT corpus along two axes, namely teacher source and composition strategy, yielding the settings in Table~\ref{tab:sft_results_presentbench_gdpval}. We consider three composition regimes. The \textbf{Single} regime uses trajectories from a single teacher--prompt source, allowing us to isolate the effect of each supervisory signal. The \textbf{Mixed} regime combines the Qwen-3.8-Max corpus with one or more GLM-5.2 variants, exposing the student to both the broad capabilities of the stronger teacher and the presentation-specific priors of the domain teacher. Building on the mixed corpora, the \textbf{Mixed+Filter} regime further introduces a lightweight outcome-driven filtering stage, which removes trajectories with rendering failures, layout overflow, content overlap, and related issues using rule-based code, thereby retaining only high-quality samples.

% ==== 放在导言区或表格前 ====
\newlength{\rowH}\setlength{\rowH}{11pt} % 约等于 \small 下每行高度
\newlength{\cmr}\setlength{\cmr}{4pt}    % 每条块内 \cmidrule 额外增加的高度(估)，用于块竖线长度
% 块竖线：#1=块内数据行数, #2=块内 \cmidrule 条数；跨整块的单条竖线(3段之一)
\newcommand{\bvline}[2]{\multirow{#1}{*}{\rule{0.4pt}{\dimexpr#1\rowH+#2\cmr-6pt\relax}}}
% Group↔Setting 段竖线(每组一段，组间留缺口)；单行组保底 6pt
\newcommand{\grule}[1]{%
  \multirow{#1}{*}{\rule{0.4pt}{\ifdim\dimexpr#1\rowH-10pt\relax<6pt 6pt\else\dimexpr#1\rowH-10pt\relax\fi}}}

\begin{table*}[t]
\centering
\small
\setlength{\tabcolsep}{4pt}
\caption{Overall SFT results on PresentBench and the GDPVal PPT subset under different teacher models and target models. For each target model, the best value in every column is bolded. Settings marked with $*$ are used as the base checkpoints for subsequent RL.}
\label{tab:sft_results_presentbench_gdpval}
\begin{tabular}{lclclcccc}
\toprule
\multirow{2}{*}{\textbf{Target Model}} & & \multirow{2}{*}{\textbf{Group}} & & \multirow{2}{*}{\textbf{Setting}} & \multicolumn{2}{c}{\textbf{PresentBench}} & \multicolumn{2}{c}{\textbf{GDPVal}} \\
\cmidrule(lr){6-7} \cmidrule(lr){8-9}
& & & & & \textbf{Score} & \textbf{$\Delta$} & \textbf{Score} & \textbf{$\Delta$} \\
\midrule
\multirow{2}{*}{Teacher} & \bvline{2}{0} & \multirow{2}{*}{--} & \grule{2}
                            & GLM-5.2       & 75.04 & -- & 74.93 & -- \\
&   &                   &   & Qwen-3.8-Max & 82.81 & -- & 82.84 & -- \\
\midrule
\multirow{11}{*}{Qwen-3.6-27B} & \bvline{11}{3}
& No Train       & \grule{1} & no train & 70.45 & -- & 74.49 & -- \\
\cmidrule(lr){5-9}
&   & \multirow{4}{*}{Single} & \grule{4}
                            & glm\_base & 70.52 & +0.07 & 71.44 & -3.05 \\
&   &               &           & glm\_pro  & 71.60 & +1.15 & 74.87 & +0.38 \\
&   &               &           & glm\_max  & 71.97 & +1.52 & 67.99 & -6.50 \\
&   &               &           & qwen      & 76.93 & +6.48 & 72.17 & -2.32 \\
\cmidrule(lr){5-9}
&   & \multirow{3}{*}{Mixed} & \grule{3}
                            & qwen+glm\_pro          & 76.66 & +6.21 & 72.71 & -1.78 \\
&   &               &           & qwen+glm\_max          & \textbf{77.90} & \textbf{+7.45} & 73.63 & -0.86 \\
&   &               &           & qwen+glm\_pro+glm\_max & 76.49 & +6.04 & 77.76 & +3.27 \\
\cmidrule(lr){5-9}
&   & \multirow{3}{*}{Mixed+Filter} & \grule{3}
                            & qwen+glm\_pro (filter) & 77.21 & +6.76 & 76.85 & +2.36 \\
&   &               &           & qwen+glm\_max (filter) & 77.59 & +7.14 & 72.00 & -2.49 \\
&   &               &           & $*$qwen+glm\_pro+glm\_max (filter) & 77.57 & +7.12 & \textbf{79.36} & \textbf{+4.87} \\
\midrule
\multirow{11}{*}{Qwen-3.6-35B-A3B} & \bvline{11}{3}
& No Train       & \grule{1} & no train & 64.93 & -- & 74.01 & -- \\
\cmidrule(lr){5-9}
&   & \multirow{4}{*}{Single} & \grule{4}
                            & glm\_base & 67.29 & +2.36 & 72.21 & -1.80 \\
&   &               &           & glm\_pro  & 68.75 & +3.82 & 70.19 & -3.82 \\
&   &               &           & glm\_max  & 68.73 & +3.80 & 73.90 & -0.11 \\
&   &               &           & qwen      & 69.55 & +4.62 & 71.68 & -2.33 \\
\cmidrule(lr){5-9}
&   & \multirow{3}{*}{Mixed} & \grule{3}
                            & qwen+glm\_pro          & 73.49 & +8.56 & 70.93 & -3.08 \\
&   &               &           & qwen+glm\_max          & 71.34 & +6.41 & 71.53 & -2.48 \\
&   &               &           & qwen+glm\_pro+glm\_max & 73.07 & +8.14 & 77.07 & +3.06 \\
\cmidrule(lr){5-9}
&   & \multirow{3}{*}{Mixed+Filter} & \grule{3}
                            & $*$qwen+glm\_pro (filter) & \textbf{74.28} & \textbf{+9.35} & \textbf{78.17} & \textbf{+4.16} \\
&   &               &           & qwen+glm\_max (filter) & 70.48 & +5.55 & 73.16 & -0.85 \\
&   &               &           & qwen+glm\_pro+glm\_max (filter) & 73.15 & +8.22 & 74.06 & +0.05 \\
\bottomrule
\end{tabular}
\end{table*}

\subsection{SFT Results}

We evaluate SFT under an in-domain/out-of-distribution protocol. As the distillation corpus is synthesized to emulate the task style of PresentBench, we use PresentBench as the in-domain benchmark and the PPT subset of GDPVal as the out-of-distribution benchmark. The results are summarized in Table~\ref{tab:sft_results_presentbench_gdpval}.

\paragraph{Our optimized SFT pipeline delivers strong in-domain gains and robust out-of-distribution generalization across different backbones.}
Across both the dense backbone (Qwen-3.6-27B) and the MoE backbone (Qwen-3.6-35B-A3B), the combination of guided distillation, teacher mixing, and quality filtering yields stable improvements over the no-SFT baselines, establishing strong same-scale performance. On PresentBench, Qwen-3.6-27B achieves a score of 77.90 ($+7.45$), surpassing the substantially larger GLM-5.2 (75.04) by 2.86 points, while Qwen-3.6-35B-A3B attains a comparable level of performance at 74.28 ($+9.35$). Notably, these improvements generalize effectively to GDPVal, where the best fine-tuned Qwen-3.6-27B and Qwen-3.6-35B-A3B reach 79.36 (\(+4.87\)) and 78.17 (\(+4.16\)), respectively, both outperforming GLM-5.2 (74.93). This demonstrates the effectiveness and generality of our SFT strategy across architectures and benchmarks.

\paragraph{Guidance strategies improve trajectory quality and strengthen supervision.}
A clear trend is that distillation data produced under stronger guidance yields better student models. Compared with the basic prompt, the guided variants deliver larger gains on both backbones, improving Qwen-3.6-27B from 70.52 to 71.97 and Qwen-3.6-35B-A3B from 67.29 to 68.73. This indicates that guidance encourages teachers to produce trajectories with stronger structural regularity and richer intermediate supervision, thereby providing more effective learning signals for the student.

\paragraph{Teacher mixing provides complementary supervision and further boosts performance.}
Mixing Qwen-3.8-Max with GLM-5.2 variants consistently improves student performance over the best single-teacher settings. On PresentBench, teacher mixing lifts Qwen-3.6-27B from 76.93 to 77.90 ($+0.97$) and Qwen-3.6-35B-A3B from 69.55 to 73.49 ($+3.94$). We attribute this gain to the synergy between the broad reasoning and planning ability of Qwen-3.8-Max and the task-specific priors of GLM-5.2 in presentation generation. Their combination exposes the student to a more diverse and informative trajectory distribution than either teacher alone.

\paragraph{Quality filtering is generally beneficial across settings.}
In most configurations, quality filtering improves student performance by mitigating noisy supervision. For example, in the mixed-teacher setting, filtering raises the PresentBench score of Qwen-3.6-35B-A3B from 73.49 to 74.28 (\(+0.79\)). At the same time, overly aggressive filtering may remove difficult yet informative samples, thereby reducing data diversity and occasionally degrading performance under certain configurations. These results highlight an important trade-off in trajectory curation, where effective distillation requires not only high-quality supervision but also adequate coverage of challenging task patterns.

\subsection{SFT Case Study}

\begin{figure*}[t]
    \centering
    \includegraphics[width=\textwidth]{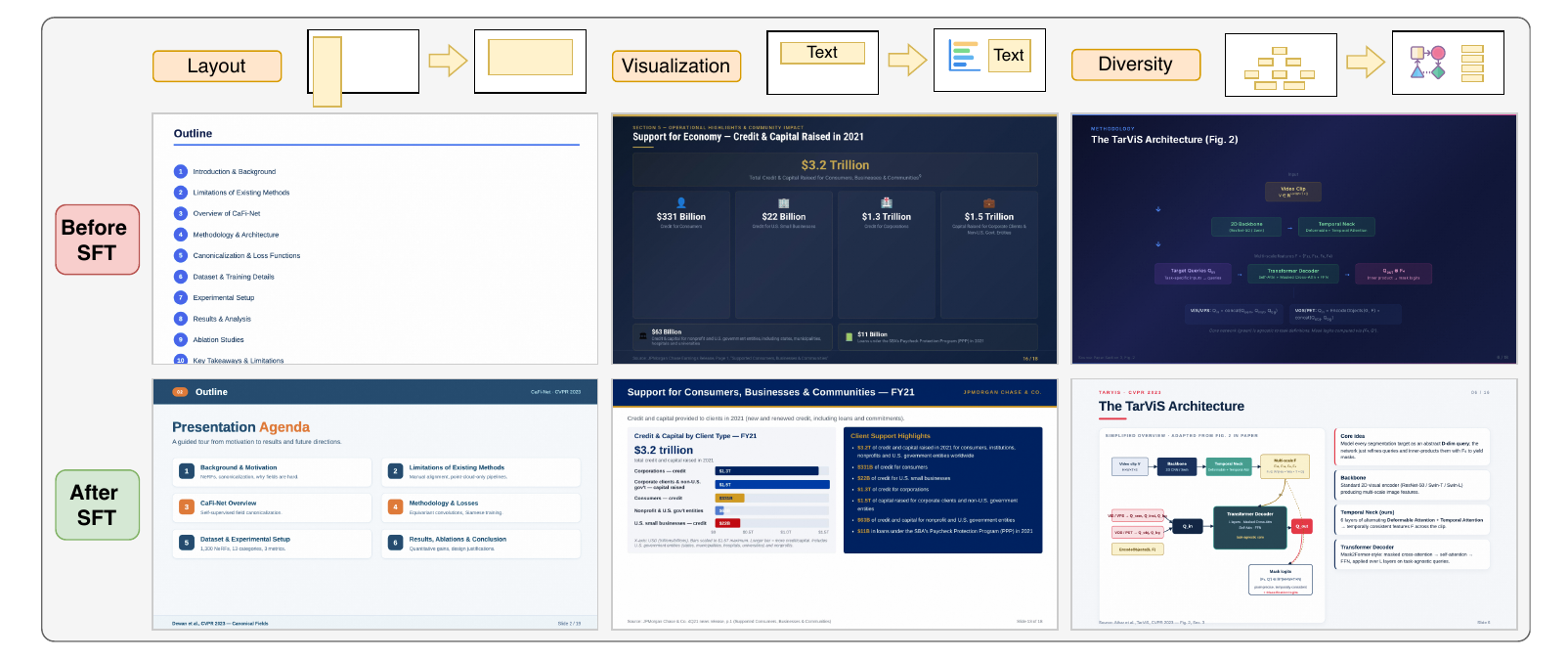}
    \caption{Case studies of slides generated before and after SFT. From left to right: Cases 1--3.}
    \label{fig:sft_case_study}
\end{figure*}

Figure~\ref{fig:sft_case_study} compares representative slides before and after SFT. Overall, SFT substantially improves both content organization and visual presentation, yielding slides with clearer hierarchy, richer visual structures, and better text--graphic coordination. These examples suggest that SFT enhances not only trajectory imitation, but also the model's internalized priors for presentation generation. However, noticeable aesthetic defects remain after SFT, including excessive whitespace, visual imbalance, and element overlap. Such issues are difficult to resolve with text-only supervision, as they require fine-grained visual feedback. This motivates our subsequent reinforcement learning stage with verifiable visual rewards to further optimize presentation aesthetics.

%% file: sections/03_4_2_rl_training.tex
\section{Verifiable RL Training}

\subsection{Initialization Model Selection}
Before constructing the RL training set, we first determine the SFT checkpoint used for policy initialization. Among the candidate SFT models, we select the checkpoint that preserves performance on the GDPVal PPT subset while attaining the highest score on PresentBench. This selection criterion is designed to ensure that RL starts from a policy that is strong in-domain without sacrificing out-of-domain generalization. Following this protocol, we adopt the corresponding post-SFT checkpoints of Qwen-3.6-27B and Qwen-3.6-35B-A3B as the dense and MoE initial policies, respectively.

\subsection{Reward Design}
To make aesthetic quality amenable to reinforcement learning, we decompose it into five verifiable dimensions defined on fully rendered slides, namely aspect-ratio compliance, whitespace adequacy, element-collision avoidance, visual balance, and font readability. Each metric is implemented as a rule-based function over the rendered output and normalized to $[0,1]$, so as to capture a distinct aspect of slide-level visual quality in a deterministic and interpretable manner. We aggregate these complementary signals with an unweighted mean to obtain the final slide-level and deck-level rewards. This verifiable reward design avoids the instability of learned critics and provides a transparent optimization objective for aesthetic refinement.

\subsection{RL Data Selection}
We construct two complementary RL datasets that capture difficulty from both task-level and aesthetic perspectives. The \textbf{task-hard dataset} is obtained by using Qwen-3.8-Max to generate presentations for all synthesized queries and selecting low-scoring instances under benchmark rubrics, thereby focusing training on globally challenging presentation-generation tasks. The \textbf{aesthetic-hard dataset} is constructed in a backbone-specific manner by running each selected SFT model on the full query pool, scoring the resulting presentations with the five verifiable aesthetic rewards, and selecting low-scoring cases that still exhibit visual deficiencies. Together, these datasets expose RL to both globally difficult queries and model-specific aesthetic failure modes, enabling targeted post-SFT refinement.

\subsection{RL with Verifiable Rewards}
\label{subsec:rlvr}

Building on the verifiable reward defined above, we further optimize the policy via an enhanced GRPO framework~\cite{shao2024deepseekmath}. To improve training stability under multi-dimensional aesthetic supervision, we incorporate several standard refinements, including reward-decoupled normalization~\cite{liu2026gdpo}, token-level optimization, and stabilized clipping~\cite{yu2026dapo}, and we study both with and without KL regularization to the SFT policy. In addition, we treat output validity as a prerequisite for reward assignment, such that only renderable presentations receive meaningful aesthetic rewards. Together, these designs establish a stable and effective RL framework for post-SFT aesthetic refinement.

At each training step, we sample 8 prompt instances, each with 8 rollouts, resulting in 64 rollouts per update for group-relative advantage estimation. Rollouts are generated with temperature 1.0. Given the long-horizon nature of slide-deck generation, each trajectory may involve multiple rounds of tool use and iterative HTML refinement, with up to 128 actions per sample. For all experiments, we train for one epoch on the dataset.  Additional training details are provided in Appendix.

% ==== 导言区需要 ====
% \usepackage{booktabs,multirow,graphicx}

\begin{table*}[t]
\centering
\footnotesize
\setlength{\tabcolsep}{4pt}
\caption{Aspect-level score matrix on PresentBench and the PPT subset of GDPVal. The best value in each column is boldfaced within each student group. Asp.\ = aspect-ratio compliance, Ws.\ = whitespace adequacy, Ele.\ = element-collision avoidance, Bal.\ = visual balance, Font = font readability, and Total = overall score.
}
\label{tab:score_matrix_all}
\resizebox{\textwidth}{!}{%
\begin{tabular}{llcccccccccccc}
\toprule
\multirow{2}{*}{\textbf{Group}} & \multirow{2}{*}{\textbf{Model}}
& \multicolumn{6}{c}{\textbf{PresentBench}} & \multicolumn{6}{c}{\textbf{GDPVal (PPT subset)}} \\
\cmidrule(lr){3-8} \cmidrule(lr){9-14}
& & \textbf{Asp.} & \textbf{Ws.} & \textbf{Ele.} & \textbf{Bal.} & \textbf{Font} & \textbf{Total}
  & \textbf{Asp.} & \textbf{Ws.} & \textbf{Ele.} & \textbf{Bal.} & \textbf{Font} & \textbf{Total} \\
\midrule
\multirow{2}{*}{Teacher}
& glm-5.2  & 0.99 & 0.87 & 0.79 & 0.54 & 0.95 & 4.13 & 0.94 & 0.84 & 0.62 & 0.50 & 0.77 & 3.67 \\
& Qwen-3.8-Max & 1.00 & 0.95 & 0.67 & 0.53 & 0.93 & 4.08 & 1.00 & 0.90 & 0.50 & 0.56 & 0.63 & 3.59 \\
\midrule
\multirow{6}{*}{Qwen-3.6-27B}
& no train                       & 0.99 & 0.76 & 0.65 & 0.48 & 0.94 & 3.81 & \textbf{0.93} & 0.69 & 0.54 & 0.43 & 0.66 & 3.26 \\
& supervised fine-tuning         & \textbf{1.00} & 0.88 & 0.74 & 0.57 & 0.95 & 4.14 & 0.70 & \textbf{0.94} & 0.57 & 0.44 & 0.70 & 3.35 \\
& rl (aesthetic-hard+KL-loss)     & 0.99 & 0.86 & 0.74 & 0.56 & 0.98 & 4.14 & 0.71 & 0.94 & 0.57 & 0.51 & 0.64 & 3.37 \\
& rl (aesthetic-hard)             & 0.99 & 0.87 & 0.75 & \textbf{0.58} & 0.97 & 4.16 & 0.76 & 0.92 & 0.48 & \textbf{0.57} & 0.71 & 3.43 \\
& rl (task-hard+KL-loss)          & 0.99 & 0.86 & 0.75 & 0.57 & 0.98 & 4.15 & 0.75 & 0.92 & 0.60 & 0.53 & \textbf{0.82} & \textbf{3.62} \\
& rl (task-hard)                  & \textbf{1.00} & \textbf{0.89} & \textbf{0.76} & 0.57 & \textbf{0.98} & \textbf{4.20} & 0.58 & 0.90 & \textbf{0.62} & 0.50 & 0.76 & 3.37 \\
\midrule
\multirow{6}{*}{Qwen-3.6-35B-A3B}
& no train                       & 0.98 & 0.80 & 0.60 & 0.50 & 0.97 & 3.85 & \textbf{0.95} & 0.78 & 0.46 & 0.51 & 0.64 & 3.34 \\
& supervised fine-tuning         & 0.99 & \textbf{0.89} & 0.74 & 0.58 & 0.95 & 4.16 & 0.88 & 0.73 & 0.49 & 0.54 & 0.80 & 3.44 \\
& rl (aesthetic-hard+KL-loss)     & 0.99 & 0.88 & 0.77 & 0.64 & 1.00 & 4.27 & 0.81 & 0.74 & 0.56 & 0.59 & \textbf{0.94} & 3.63 \\
& rl (aesthetic-hard)             & \textbf{1.00} & 0.84 & \textbf{0.86} & 0.57 & 0.99 & 4.27 & 0.87 & 0.83 & 0.68 & 0.50 & 0.87 & 3.74 \\
& rl (task-hard+KL-loss)          & \textbf{1.00} & 0.85 & 0.81 & 0.64 & 1.00 & 4.29 & 0.74 & \textbf{0.88} & 0.63 & \textbf{0.59} & 0.87 & 3.71 \\
& rl (task-hard)                  & \textbf{1.00} & 0.87 & 0.85 & \textbf{0.64} & \textbf{1.00} & \textbf{4.36} & 0.83 & 0.78 & \textbf{0.75} & 0.52 & 0.87 & \textbf{3.75} \\
\bottomrule
\end{tabular}%
}
\end{table*}

%% file: sections/03_4_3_rl_result.tex
\subsection{RL Result}

\begin{figure*}[t]
    \centering
    \includegraphics[width=\textwidth]{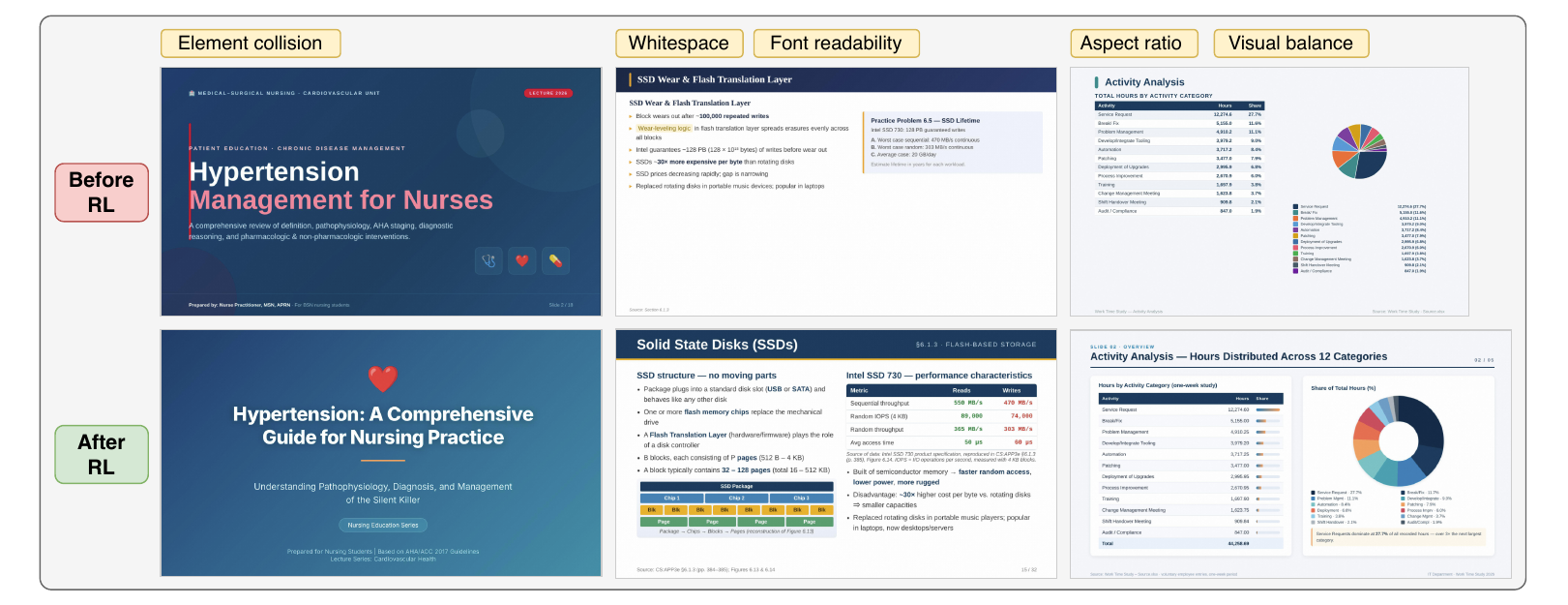}
    \caption{Case studies of slides generated before and after RL. From left to right, three cases exhibit clear improvements across five aesthetic metrics.}
    \label{fig:rl_case_study}
\end{figure*}

\paragraph{Main Results}
Table~\ref{tab:score_matrix_all} reports model performance on the five verifiable aesthetic dimensions. The three examples in Figure~\ref{fig:rl_case_study} offer a more concrete view of how RL improves the five aesthetic dimensions. Reinforcement learning improves aesthetic quality consistently across all five metrics, with especially pronounced gains on the MoE backbone. In particular, the best Qwen-3.6-35B-A3B variant achieves overall scores of 4.36 on PresentBench and 3.75 on the GDPVal PPT subset, outperforming both the no-train and SFT baselines. Qwen-3.6-27B also benefits from RL, though the improvements are less uniform across settings. We attribute the stronger gains of the MoE backbone to its higher output diversity, which induces richer reward variation, and to expert specialization, which enables aesthetic updates to be routed more effectively than in dense models. Overall, these results show that our pipeline yields progressive gains in slide aesthetics from no training to SFT and further to RL.

\paragraph{Ablation on data selection.}
Task-hard consistently outperforms aesthetic-hard on both backbones. We attribute this advantage to its broader coverage of content and layout challenges, which induces greater reward variation and more informative policy updates. By contrast, aesthetic-hard targets a narrower subset of failure modes and is therefore more susceptible to weak advantage signals and local overfitting. This finding provides a useful insight for RL data selection, where broad difficulty coverage maybe proves more effective than narrowly targeted sampling.

\paragraph{Ablation on KL divergence.}
On PresentBench, removing KL regularization is consistently equal to or better than retaining it under the same data-selection strategy. Since our reward is fully verifiable and invalid outputs are excluded by the render-validity constraint, KL mainly acts as a conservative prior that restricts movement toward higher-reward layouts. Removing it therefore enables more effective in-domain aesthetic optimization.

%% file: sections/06_conclusion.tex
\section{Conclusion}

In this paper, we investigate adaptation-oriented training methods for automatic PPT generation. To identify the most promising generation paradigm, we first survey and systematically compare mainstream approaches to automatic presentation generation. Our analysis further reveals that existing training paradigms remain relatively rudimentary, lacking effective data synthesis pipelines, high-quality teacher-guidance strategies, and end-to-end RL designs tailored to this task. To address these limitations, we propose a unified training framework that integrates cross-validated data synthesis, guided teacher enhancement, and reinforcement learning with verifiable rewards. Specifically, the framework first constructs a cross-validated pipeline for synthesizing valid and challenging training instances, then improves supervised fine-tuning through teacher enhancement, teacher mixing, and lightweight outcome-driven trajectory filtering, and finally introduces verifiable aesthetic metrics within a GRPO-based RL framework to directly optimize slide-level visual quality. Extensive experiments show that our method consistently improves both dense and MoE backbones, with SFT substantially enhancing overall generation quality and RL further refining multiple aesthetic dimensions beyond what text-only supervision can achieve. Overall, our framework offers an effective path toward more capable and reliable presentation intelligence.

%% file: sections/08_AI_use_statement.tex
\section*{AI Use Statement}

In this work, we used generative AI tools for generating synthetic datasets, 
cleaning and reformatting datasets, and assisting with translation. 
We have not used generative AI tools for other tasks with required disclosure, 
and the remaining required disclosure tasks are not applicable to this work. 
We have reviewed all AI-assisted work and verified the quality and correctness 
of the resulting data and translated content. We take responsibility for the 
final content of this work, including text, claims, and artifacts produced with the aid of generative AI.

%% file: sections/07_00_appendix_method_selection.tex
\appendix

\section{Details of Presentation Generation Method Selection}

Existing mainstream LLM-based approaches to automated presentation generation can be broadly categorized into three paradigms: Python-based code generation and execution, HTML-based rendering, and direct image generation. Representative systems, such as AutoPresent~\cite{ge2025autopresent} and Talk-to-Your-Slides~\cite{jung2025talk}, synthesize executable Python code to generate presentation files in the \texttt{.pptx} format. In contrast, methods such as AeSlides~\cite{pan2026aeslides} and PreGenie~\cite{xu2025pregenie} cast presentation generation as an HTML synthesis and rendering pipeline. Beyond these structured generation paradigms, recent benchmarks, e.g., SlideBench~\cite{ge2025autopresent}, have further explored end-to-end image-based generation using advanced generative models such as Stable Diffusion~\cite{rombach2022high} and DALL·E~3~\cite{betker2023improving}. These paradigms differ substantially in representational flexibility, controllability, and system extensibility, thereby exhibiting distinct advantages and inherent limitations.

Through extensive investigation and empirical analysis, we observe that presentations generated via the image-based paradigm frequently suffer from textual hallucinations, including non-existent Chinese characters and misspelled English words. Furthermore, this paradigm is characterized by higher instability, limited controllability, and greater computational costs. Conversely, the continuously advancing code generation capabilities of LLMs demonstrate significantly greater potential for reliable presentation synthesis. Consequently, our study primarily focuses on evaluating the efficacy of LLMs in generating presentations through two code-centric paradigms.

\paragraph{Framework Selection.}
Given the inherent complexity of presentation generation—which encompasses diverse output formats, multifaceted content requirements, and heterogeneous cognitive tasks such as document comprehension, information retrieval, and content drafting—we adopt OpenClaw as our comprehensive agentic framework for experimentation. The rationale for this selection is threefold:
\begin{itemize}
    \item OpenClaw operates on a closed-loop agent architecture comprising ``planning--tool invocation--observation--reflection--iteration,'' which aligns naturally with the iterative reasoning process required for presentation generation.
    \item It provides a comprehensive toolset and robust agent orchestration mechanisms, enabling high-quality execution across diverse task types and demonstrating strong generalizability to varying presentation generation requirements.
    \item It natively supports multiple generation paradigms; a single model can drive both Python-based and HTML-based generation without any framework-level modifications, thereby ensuring fair and consistent evaluation across paradigms.
\end{itemize}

\paragraph{Evaluation Setup.}
We employ OpenClaw powered by Qwen-3.6-27B and Qwen-3.6-35B-A3B to conduct experiments across four presentation generation configurations:
\begin{itemize}
    \item Single-page HTML: Generates multiple HTML files, each containing the content of a single slide.
    \item Multi-page HTML: Generates a single HTML file encompassing the entire presentation.
    \item Single-page Python: Generates multiple Python scripts, each responsible for producing one slide.
    \item Multi-page Python: Generates a single Python script that produces the complete presentation.
\end{itemize}

Evaluation is conducted on the full PresentBench benchmark~\cite{chen2026presentbench} and the presentation generation subset of GDPVal~\cite{patwardhan2026gdpval}. The generated HTML files are converted into rasterized PDFs by capturing each slide via browser-based screenshots and then merging them, thereby ensuring faithful preservation of the actual on-screen rendering. The Python scripts are executed within the task pipeline to generate presentation files, which are subsequently converted into PDF automatically using headless LibreOffice. All results are evaluated in a unified manner by prompting Gemini-3-Flash-Preview~\cite{google2026gemini3flashpreview} to score each PDF independently on each predefined rubric, following the protocol specified by the corresponding benchmark. This per-rubric evaluation paradigm mitigates the degradation in assessment accuracy caused by jointly scoring multiple criteria, such as excessive context length and cross-rubric interference.

\paragraph{Evaluation Results.}

\begin{table}[t]
\centering
\small
\setlength{\tabcolsep}{3pt}
\caption{Results of Qwen-3.6-27B on PresentBench under four generation settings.}
\label{tab:presentbench_qwen27b_all}
\begin{tabular}{llcccc}
\toprule
\textbf{Section} & \textbf{Metric} & \textbf{Single-HTML} & \textbf{Multi-HTML} & \textbf{Single-Python} & \textbf{Multi-Python} \\
\midrule
\multirow{5}{*}{Category} 
& Academia    & 74.73 & 65.31 & 62.79 & 62.52 \\
& Advertising & 54.91 & 53.30 & 49.99 & 48.94 \\
& Economics   & 64.73 & 60.12 & 52.77 & 55.33 \\
& Education   & 71.59 & 68.73 & 61.17 & 63.91 \\
& Talk        & 71.26 & 56.75 & 60.89 & 63.30 \\
\midrule
\multirow{5}{*}{Dimension (\%)} 
& Content Quality        & 91.92 & 85.85 & 83.32 & 83.28 \\
& Visual Design          & 60.18 & 51.72 & 28.77 & 31.37 \\
& Content Completeness   & 76.42 & 70.35 & 73.92 & 75.37 \\
& Content Correctness    & 66.85 & 58.48 & 61.11 & 62.09 \\
& Slide-wise Consistency & 54.93 & 48.69 & 49.35 & 50.82 \\
\midrule
\multirow{1}{*}{Overall} 
& Average Score & \textbf{70.45} & 63.39 & 59.56 & 60.80 \\
\bottomrule
\end{tabular}
\end{table}

\begin{table}[t]
\centering
\small
\setlength{\tabcolsep}{3pt}
\caption{Results of Qwen-3.6-35B-A3B on PresentBench under four generation settings.}
\label{tab:presentbench_qwen35b_all}
\begin{tabular}{llcccc}
\toprule
\textbf{Section} & \textbf{Metric} & \textbf{Single-HTML} & \textbf{Multi-HTML} & \textbf{Single-Python} & \textbf{Multi-Python} \\
\midrule
\multirow{5}{*}{Category} 
& Academia    & 67.72 & 69.43 & 68.02 & 67.93 \\
& Advertising & 53.50 & 55.48 & 48.75 & 54.06 \\
& Economics   & 58.00 & 59.63 & 57.54 & 58.86 \\
& Education   & 66.58 & 65.86 & 66.51 & 66.82 \\
& Talk        & 68.72 & 56.56 & 64.56 & 65.96 \\
\midrule
\multirow{5}{*}{Dimension (\%)} 
& Content Quality        & 87.56 & 84.88 & 86.72 & 87.44 \\
& Visual Design          & 51.31 & 54.65 & 37.75 & 38.01 \\
& Content Completeness   & 74.57 & 69.59 & 76.03 & 75.75 \\
& Content Correctness    & 60.95 & 60.11 & 65.37 & 67.21 \\
& Slide-wise Consistency & 47.89 & 51.09 & 53.27 & 55.08 \\
\midrule
\multirow{1}{*}{Overall} 
& Average Score & \textbf{64.93} & 64.28 & 64.10 & 64.91 \\
\bottomrule
\end{tabular}
\end{table}

For the PresentBench dataset, we report the overall average score alongside category-level and dimension-level performance. The evaluation criteria for each dimension are defined as follows: (1) Content Quality: measures the overall quality of the presentation content, including slide quantity, logical coherence, and clarity of key points. (2) Visual Design: evaluates layout, typography, color composition, and image usage. (3) Content Completeness: assesses whether the presentation adequately covers the key points of the source material. (4) Content Correctness: measures the factual accuracy of data, claims, and citations. (5) Slide-wise Consistency: evaluates whether each slide remains aligned with the corresponding source material.

As shown in Tables~\ref{tab:presentbench_qwen27b_all} and~\ref{tab:presentbench_qwen35b_all}, the Single-HTML generation paradigm consistently attains the best overall performance for both Qwen-3.6-27B and Qwen-3.6-35B-A3B. Furthermore, on the visual-design dimension, HTML-based generation markedly outperforms Python-based generation across both models. Through qualitative analysis, we find that HTML offers substantial advantages over Python in several aspects that are crucial to high-quality presentation synthesis, including layout composition (e.g., overall structural organization and control of visual centering), color design (e.g., coordinated background--text palettes and gradient effects), visual diversity (e.g., the flexible incorporation of fonts, icons, and other stylistic elements), and mathematical expression rendering. In particular, since python-pptx does not provide a native mathematical typesetting engine, formulas can only be rendered as plain text, which significantly constrains expressive fidelity in presentations requiring formal notation.

A further comparison between the two HTML-based paradigms suggests that Single-HTML is more advantageous than Multi-HTML in terms of robustness and optimization efficiency. Specifically, slide-wise independent generation naturally supports fault isolation, precise error localization, and incremental refinement. By contrast, the Multi-HTML paradigm requires the model to jointly maintain all slides in the deck together with global CSS within a single file, which more easily introduces failure modes such as blank pages and rendering artifacts. Moreover, as shown in Table~\ref{tab:gdpval_presentbench_avg}, the same conclusion is consistently reproduced on the presentation generation subset of GDPVal. Based on these empirical observations and qualitative findings, we adopt Single-HTML as the presentation generation paradigm for subsequent training.

%% file: sections/07_01_appendix_data_synthesis.tex
\section{Details of Cross-Validated Data Synthesis}

\subsection{Data Synthesis Pipeline}
We design a five-stage pipeline for high-quality data synthesis for PPT generation, comprising source profiling, task construction, rubric generation, principle audit, and rubric tightening.

\paragraph{Source profiling.}
We begin with a large and diverse document pool and retain only sources that are genuinely suitable for presentation generation. A useful source should not merely be readable; it should also contain sufficient structure and evidence to support both slide planning and downstream evaluation. For each admitted document, we construct a structured source card that summarizes its major sections, key quantities, figures, domain terms, and common factual pitfalls. This representation provides subsequent stages with a clean and standardized abstraction of the source material, while also supplying the reference anchors against which later tasks and rubrics are cross-checked.

\paragraph{Task construction.}
We then transform each source card into a slide-generation task under a cross-validated construction rule. The key challenge at this stage is to prevent answer leakage. If the instruction already reveals the facts that will later be judged, the task collapses into template completion and no longer evaluates genuine understanding. To avoid this failure mode, task construction is constrained by information that is deliberately incomplete with respect to rubric construction. Specifically, source-card content is partitioned into scope fields and answer fields. Scope fields are used to define what the presentation should cover, while answer fields are withheld from the task and reserved exclusively for later verification. As a result, the task is validated not by how specific it appears in isolation, but by whether it remains informative without exposing the evidence required for scoring. In this sense, the legality of the task is defined externally by its non-overlap with the rubric's answer space.

\paragraph{Rubric generation.}
We next synthesize an instance-level rubric by cross-validating it against both the source card and the task. Each rubric contains two binary checklists, one for completeness and one for correctness. The completeness checklist verifies whether the task-required content is present, whereas the correctness checklist verifies whether that content is faithful to the source. This separation ensures that the rubric does not merely restate the task, but checks it from a different evidential view. To avoid rubrics that are either too sparse or insufficiently demanding, we scale rubric size with source information density rather than a fixed template. Moreover, every key quantity and every factual trap extracted from the source card must be covered by at least one rubric item. The rubric is therefore not accepted because it looks well formed on its own, but because it can be externally grounded in source evidence while remaining complementary to the task specification.

\paragraph{Principle audit.}
After task and rubric synthesis, we conduct an explicit audit to verify that the resulting pair satisfies our design principles. The audit examines, among other aspects, instruction format, rubric specificity, the separation between completeness and correctness, and item atomicity. More importantly, it checks the consistency of the task--rubric pair as a coupled object rather than validating either artifact in isolation. Cases that fail critical checks are routed to a focused repair step and subsequently re-audited. The purpose of this stage is not to determine whether the rubric is already sufficiently challenging in practice, but to ensure that the synthesized task and rubric satisfy the structural requirements imposed by their cross-validated construction.

\paragraph{Rubric tightening.}
Even structurally valid rubrics may remain too permissive for strong generation models. We therefore introduce a final tightening stage to restore discriminative power. The objective is to make existing criteria more stringent without altering the underlying measurement target. To this end, we keep the overall rubric structure fixed, including dimensions, weights, and item counts, and only strengthen the wording of individual checklist items. Because the rubric shape is preserved, changes in score can be attributed to stricter evaluation rather than to metric redefinition. This final stage extends the same cross-validated philosophy to difficulty calibration, ensuring that rubric refinement strengthens evaluation without breaking comparability.

Overall, the proposed pipeline provides a scalable approach to synthesizing presentation tasks and rubrics without human-in-the-loop authoring. More importantly, it makes answer non-disclosure, source-grounded coverage, and evaluation difficulty explicit cross-stage design objectives rather than informal prompt preferences, thereby providing a stronger foundation for both supervised fine-tuning and verifiable reward optimization.

\subsection{Domain Taxonomy}
Table~\ref{tab:unified_domain_taxonomy_en} presents the complete unified domain taxonomy used in our data synthesis pipeline. The taxonomy covers 25 top-level domains and 162 subdomains.

\begin{longtable}{p{0.05\textwidth} p{0.22\textwidth} p{0.68\textwidth}}
\caption{Complete domain taxonomy used in the data synthesis pipeline.}
\label{tab:unified_domain_taxonomy_en} \\
\toprule
\textbf{\#} & \textbf{Top-level Domain} & \textbf{Subdomains} \\
\midrule
\endfirsthead

\toprule
\textbf{\#} & \textbf{Top-level Domain} & \textbf{Subdomains} \\
\midrule
\endhead

\midrule
\multicolumn{3}{r}{Continued on next page} \\
\endfoot

\bottomrule
\endlastfoot

1 & Technology \& Internet & Product introduction; Technical solution briefing; Architecture design; ERP \& IT systems; Data analysis report; Technology trend insight; Fundraising pitch deck; Software engineering; Algorithms \& AI engineering; IT operations \& DevOps; Cybersecurity; Product requirements \& roadmap \\ \midrule

2 & Healthcare & Clinical case sharing; Medical lecture; Pharmaceutical introduction; Health management; Clinical guideline interpretation; Medical device introduction \\ \midrule

3 & Legal \& Compliance & Legal training; Contract interpretation; Policy briefing; Mediation; Intellectual property; Compliance, internal control \& regulatory filing; Data compliance \& privacy \\ \midrule

4 & Government \& Public Sector & Government work report; Policy interpretation; Party education session; Workplace safety; Digital government; Rural revitalisation briefing \\ \midrule

5 & Real Estate \& Construction & Project proposal; Design review; Completion report; Urban renewal; Property marketing; Construction progress report \\ \midrule

6 & Automotive Manufacturing & New vehicle launch; Supply chain analysis; Quality report; Assembly process; Connected-vehicle solution; Smart manufacturing \& IIoT \\ \midrule

7 & Travel, Culture \& Food Service & Attraction promotion; Franchise proposal; Event plan; Cooking recipe; Tour itinerary planning; Hotel \& lodging operations \\ \midrule

8 & HR \& Administration & New-hire onboarding; Policy rollout; Annual review; Monthly \& weekly report; Recruiting \& staffing; Performance \& compensation; Culture \& team-building plan \\ \midrule

9 & E-commerce \& Retail & Sales-event retrospective; Category analysis; Merchant recruitment roadshow; Apparel \& textiles; Live-commerce operations; Membership \& private-traffic operations \\ \midrule

10 & Environment \& Energy & Carbon neutrality; Renewable energy; Energy saving \& emission reduction; Seawater desalination; Environmental impact assessment; ESG \& sustainability report \\ \midrule

11 & Telecom \& Electronics & Communication networks; Signal processing; Electronic circuits; Base stations; 5G private-network solution; IoT solution; Chips \& semiconductors \\ \midrule

12 & Literature, Humanities \& History & Classical literature; Historical events; Philosophy \& religion; Language-arts teaching; Cultural heritage interpretation; Book-club presentation \\ \midrule

13 & Sports & Event planning; Sports \& health; Club proposal; Fitness training course; Sports marketing \& sponsorship \\ \midrule

14 & Agriculture \& Food & Cultivation technique; Food safety; Agricultural product promotion; Tobacco; Food R\&D \& new products; Smart agriculture solution \\ \midrule

15 & Personal Life & Birthday celebration; Wedding \& anniversary; Personal resume; Parenting; Travel guide; Household financial planning; Personal development plan \\ \midrule

16 & Finance \& Accounting & Financial analysis \& reconciliation; Budgeting \& cost management; Audit \& internal control; Investment, financing \& valuation; Financial risk \& insurance; Earnings interpretation \& results release; Earnings release; Interim financial report; Quarterly financial report; Shareholder update \& quarterly letter; Macro flagship report \\ \midrule

17 & Sales \& Marketing & Marketing campaign plan; Brand \& advertising; Sales performance review; Market \& customer research; Sales training \& methodology; Channel \& distributor conference \\ \midrule

18 & Operations \& Supply Chain & Procurement \& supplier management; Inventory \& logistics; Production \& operations planning; Quality management; Lean operations \& process improvement \\ \midrule

19 & Strategy Consulting & Strategic planning; Industry research report; Competitive analysis; Business plan \& feasibility study; Digital transformation plan; M\&A integration plan; Macroeconomic \& policy research \\ \midrule

20 & Project Management & Project plan \& milestones; Risk \& change management; Project status report; Project retrospective \& PMO; Project initiation \& kickoff; Tender \& bidding proposal \\ \midrule

21 & Customer Service & Service-standard training; Complaint \& dispute handling; Ticketing \& SLA management; Customer success \& FAQ; Customer-experience improvement \\ \midrule

22 & Media \& Journalism & Story selection \& reporting; Fact checking; Investigative reporting; Public-sentiment monitoring; Brand \& crisis communications; New-media content operations \\ \midrule

23 & Education \& Training & Instructional design; Curriculum development; Learning analytics \& assessment; Academic administration; Admissions presentation; Demonstration lesson \\ \midrule

24 & Academic Research & Research paper presentation; Conference talk; Proposal \& thesis defence; Literature review; Grant proposal; Natural science research; Social science research \\ \midrule

25 & Entertainment \& Creative & Film \& video production; Game development; Design \& creative; Animation \& IP operations \\

\bottomrule
\end{longtable}

%% file: sections/07_02_appendix_prompt.tex
\section{Details of Guided Supervised Fine-Tuning }
% \subsection{Distillation Prompt Templates}
% \label{app:prompts}

We instantiate three prompt templates of increasing guidance strength. Each is
prepended to the task-specific instruction file (\texttt{instructions.md}, denoted
below by the placeholder \texttt{<<< TASK INSTRUCTIONS (instructions.md) >>>})
and followed by a fixed \emph{Expected Output} block. The \textbf{base} template
adds no guidance beyond the task and output format; the \textbf{pro} template
prepends a flat checklist of fidelity and layout requirements; and the
\textbf{max} template organizes the guidance into a structured methodology with a
mandatory render--inspect--fix verification loop.

\label{app:prompt-base}
\begin{lstlisting}[
  style=prompt,
  title={\bfseries Base Prompt}
]
<<< TASK INSTRUCTIONS (instructions.md) >>>

Expected Output:
A **complete slide deck** satisfying all constraints above.
A folder `html_slides/` containing one HTML file per slide
(slide_1.html, slide_2.html, ...).
\end{lstlisting}

\label{app:prompt-pro}
\begin{lstlisting}[
  style=prompt,
  title={\bfseries Pro Prompt}
]

**IMPORTANT: You must generate the presentation as separate HTML files -- one file per slide -- NOT as a PowerPoint (.pptx) file or a single combined HTML file.**

Create a folder called `html_slides/` under `/workspace/` and generate one self-contained HTML file per slide, named `slide_1.html`, `slide_2.html`, ..., `slide_{n}.html`. Each HTML file represents exactly one slide and must be fully self-contained (all styles inline) and viewable by opening it directly in a browser. Use professional styling with proper fonts, colors, and layout to make each slide look like a real presentation slide.

---

**CRITICAL -- You MUST read any reference materials before creating slides:**

Before you start generating slides, check if there are any reference files placed under `/workspace/` (e.g., PDFs, Markdown files, text files, spreadsheets, images, etc.). If reference materials exist, you **MUST** read and thoroughly analyze them first. The task description provided below may only be a brief summary -- it is NOT sufficient to create an accurate, high-quality presentation on its own. You must extract detailed content, exact numbers, data, and relevant information directly from the reference materials. **Do NOT rely solely on the text instructions below.** Any content that cannot be traced back to the reference materials (when they exist) may be considered fabricated and will receive a low score.

---
**IMPORTANT: PPT generation must strictly follow the source material, and no information may be fabricated. You must adhere to the following principles:**

1. **Page Count Lock**: Generate the exact number of slides requested and output a page-by-page plan before starting.
2. **Per-Sentence Traceability**: Every factual statement must map to an explicit supporting sentence in the source; delete anything that cannot be traced.
3. **Zero Numerical Tolerance**: Compare every table cell against the source and run a reverse diff after generation; no rounding or approximation.
4. **Term Fidelity**: Copy loss names, module names, and variable symbols verbatim from the source; never rewrite or invent new names.
5. **Symbol-Level Formula Check**: Verify every LaTeX formula symbol-by-symbol against the source equation, including subscripts, norms, operators, and distance definitions.
6. **Formal Definition Priority**: When the source provides a formal definition, copy it exactly; never simplify by altering summation ranges or omitting subscripts.
7. **Bullet Count Cap**: No block on any slide may exceed 6 bullets; split into multiple slides or convert to a table when over.
8. **Minimum Font Size**: Enforce strict lower bounds -- table cells >= 14pt, body text >= 18pt, sub-headings >= 22pt, slide titles >= 28pt, deck title >= 36pt, and chart axis labels >= 12pt -- to ensure readability from the back row at 1080p projection scale.
9. **Zero Overlap**: Keep footer citations clear of tables and figures with adequate padding; verify via bounding-box checks after generation.
10. **Capitalization Consistency**: Pick one header style for the entire deck and apply it uniformly, including section-marker conventions.
11. **Letter-Spacing Check**: Forbid titles broken into spaced-out characters; re-typeset whenever font subsetting artifacts appear.
12. **Chart Axis Labeling**: All charts must carry labeled axes with units, and qualitative charts must annotate the better/worse direction.
13. **Visual-Semantic Alignment**: Bar lengths and visual encodings must match the underlying metric direction so that "lower-is-better" and "higher-is-better" stay consistent with the labels.
14. **Spelling Whitelist**: Lock high-risk domain terms to their canonical spellings and reject any near-form drift variants.
15. **English Grammar**: Grammar must be correct with no spelling errors; forbid run-on sentences, verb/noun confusions, and incorrect plural attributive nouns in compound terms.

---
**CRITICAL -- The generated PPT needs to satisfy the following requirements:

1. The aspect ratio must follow the standard 16:9 format. Set each HTML page size to 1280px wide and 720px tall.
2. Except for the title slide, table of contents slide, and ending slide, other slides should not contain excessive white space.
3. Prevent element collisions, including overlapping elements, overflowing their parent containers, or extending beyond the slide boundaries.
4. The visual composition should remain as balanced as possible, and the visual center should not be overly shifted away from the center of the canvas.
5. The PPT should include rich visual elements such as icons, images, and charts.
6. Avoid dense, long paragraphs of small text, as they negatively affect visual readability. Except for footnotes and similar auxiliary content, all other text must use a font size of at least 14pt.

<<< TASK INSTRUCTIONS (instructions.md) >>>

Expected Output:
A **complete slide deck** satisfying all constraints above.
A folder `html_slides/` containing one HTML file per slide (slide_1.html, slide_2.html, ...)
\end{lstlisting}

\label{app:prompt-max}
\begin{lstlisting}[
  style=prompt,
  title={\bfseries Max Prompt}
]

**IMPORTANT: You must generate the presentation as separate HTML files -- one file per slide -- NOT as a PowerPoint (.pptx) file or a single combined HTML file.**

Create a folder called `html_slides/` under `/workspace/` and generate one self-contained HTML file per slide, named `slide_1.html`, `slide_2.html`, ..., `slide_{n}.html`. Each HTML file represents exactly one slide and must be fully self-contained (all styles inline, icons/emblems/charts as inline SVG or CSS shapes) and viewable by opening it directly in a browser with no external assets.

---

## 1. Read Reference Materials FIRST

Before generating any slides, check if there are reference files under `/workspace/` (e.g., PDFs, Markdown, text, spreadsheets, images). If they exist, you **MUST** read and thoroughly analyze them first. The task description below may only be a brief summary -- it is NOT sufficient alone. Extract detailed content, exact numbers, data, and relevant information directly from the reference materials. **Do NOT rely solely on the text instructions.** Any content that cannot be traced back to the reference materials (when they exist) may be considered fabricated and will receive a low score.

---

## 2. Establish a Design System BEFORE Writing Any Slide

- **Canvas**: every slide is exactly `width:1280px; height:720px; position:relative; overflow:hidden` (standard 16:9).
- Define ONE reusable palette (background, primary, accent, text colors) and ONE typographic scale; apply them identically on every slide (consistent header, footer, page number, section tag).
- **Font size lower bounds** (strictly enforced):
  - Deck title >= 36pt
  - Slide titles >= 28pt
  - Sub-headings >= 22pt
  - Body/bullet text >= 18pt
  - Table cells >= 14pt
  - Chart axis labels >= 12pt
  - Nothing may go below 12pt.
- Pick one header capitalization style for the entire deck and apply it uniformly.

---

## 3. Source Fidelity -- Zero Fabrication

1. **Page Count Lock**: Generate the exact number of slides requested; output a page-by-page plan before starting.
2. **Per-Sentence Traceability**: Every factual statement must map to an explicit supporting sentence in the source; delete anything untraceable.
3. **Zero Numerical Tolerance**: Compare every table cell against the source and run a reverse diff after generation; no rounding or approximation.
4. **Term Fidelity**: Copy loss names, module names, and variable symbols verbatim from the source; never rewrite or invent new names.
5. **Symbol-Level Formula Check**: Verify every LaTeX formula symbol-by-symbol against the source equation, including subscripts, norms, operators, and distance definitions.
6. **Formal Definition Priority**: When the source provides a formal definition, copy it exactly; never simplify by altering summation ranges or omitting subscripts.
7. **Spelling Whitelist**: Lock high-risk domain terms to their canonical spellings; reject any near-form drift variants.
8. **English Grammar**: Grammar must be correct with no spelling errors; forbid run-on sentences, verb/noun confusions, and incorrect plural attributive nouns.

---

## 4. Layout & Composition Rules

- **Usable content height** = 720 - header - footer - vertical paddings. Compute it explicitly before placing content.
- Before finalizing a slide, estimate the stacked height of all blocks (title + boxes + bullets x line-height) and confirm it fits with margin to spare.
- Keep a safe inner margin from all four edges; nothing may touch or cross slide borders.
- **Balanced composition**: distribute content so visual mass is centered -- no large empty quadrant, no crowded corner. If one side is heavy, add a supporting panel, chart, stat, or graphic to balance it. Prefer CSS grid/flex with justify/align centering over free-floating elements.
- **Bullet Count Cap**: No block on any slide may exceed 6 bullets; split into multiple slides or convert to a table when over.
- **Zero Overlap**: Keep footer citations clear of tables and figures with adequate padding; verify via bounding-box checks.
- **No excessive white space** (except title, TOC, and ending slides).
- **Rich visual elements**: include icons, images, charts, and diagrams where appropriate.
- **Chart requirements**: all charts must carry labeled axes with units; qualitative charts must annotate better/worse direction; bar lengths and visual encodings must match the underlying metric direction.

---

## 5. MANDATORY Render -> Inspect -> Fix Loop (EVERY Slide)

After writing each slide, render it to a screenshot and visually inspect it. Check for:
  (a) element overlap,
  (b) text/box overflow or content cut off at edges,
  (c) fonts rendering too small,
  (d) empty or unbalanced regions,
  (e) footer collision with other elements,
  (f) titles broken into spaced-out characters (font subsetting artifacts).

Fix every issue with targeted edits, then re-render to confirm. **Repeat until clean. Do NOT consider a slide done until a screenshot proves it has no overlap, overflow, or layout violations.**

---

Deliverable: N clean, visually balanced, overflow-free, source-faithful slides that share one consistent design system, each verified by screenshot.

<<< TASK INSTRUCTIONS (instructions.md) >>>

Expected Output:
A **complete slide deck** satisfying all constraints above.
A folder `html_slides/` containing one HTML file per slide (slide_1.html, slide_2.html, ...)
\end{lstlisting}

%% file: sections/07_03_appendix_sft.tex
\section{Supervised Fine-Tuning Implementation Details}

\subsection{Training Details}

All SFT runs are built on a Megatron-based training stack (Turbo) and executed on 64 GPUs. We adopt full-parameter fine-tuning with bf16 and thinking mode enabled. Optimization uses the Adam optimizer wrapped in a Megatron distributed optimizer, with peak learning rate 5e-6, cosine decay, weight decay 0.01, gradient clipping 1.0, and per-device batch size 1. We use a micro-batch size of 1 per data-parallel replica and gradient accumulation of 4. The maximum context length is 131,072 tokens. For memory efficiency, we enable full activation recomputation (uniform method, 1 layer per checkpoint), sequence parallelism, gradient-reduce overlap, the distributed optimizer, and the TransformerEngine backend with fused bias-activation; the allocator is configured with expandable\_segments and NVLS is disabled. 
The two backbones use different parallelism layouts. The dense Qwen-3.6-27B uses tensor-parallel = 2, context-parallel = 8, and pipeline-parallel = 4 (without expert parallelism), yielding data-parallel = 1 over the 64 GPUs. The MoE Qwen-3.6-35B-A3B uses tensor-parallel = 2, context-parallel = 4, pipeline-parallel = 4, and expert-parallel = 2, also yielding data-parallel = 1. For the MoE model, experts are dispatched with an all-to-all token dispatcher and grouped GEMM, an expert capacity factor of 1.0 with padding to capacity, a probability-based token-drop policy, and shared-expert overlap disabled. Additionally, we use a warmup ratio of 1/10 for the 27B model and 1/20 for the 35B-A3B model.

\begin{table*}[t]
\centering
\small
\setlength{\tabcolsep}{2.5pt}
\caption{Per-module PresentBench scores for two target models (Qwen-3.6-27B and Qwen-3.6-35B-A3B) under different teacher references and supervised fine-tuning configurations. Module abbreviations are as follows: Acad.\ = Academia, Adv.\ = Advertising, Econ.\ = Economics, Edu.\ = Education, and Talk = Talk.}
\label{tab:sft_per_module}
\begin{tabular}{lclclccccc}
\toprule
\multirow{2}{*}{\textbf{Target Model}} & & \multirow{2}{*}{\textbf{Group}} & & \multirow{2}{*}{\textbf{Setting}} & \multicolumn{5}{c}{\textbf{PresentBench Modules}} \\
\cmidrule(lr){6-10}
& & & & & \textbf{Acad.} & \textbf{Adv.} & \textbf{Econ.} & \textbf{Edu.} & \textbf{Talk} \\
\midrule
\multirow{2}{*}{Teacher} & \bvline{2}{0} & \multirow{2}{*}{--} & \grule{2}
                            & GLM-5.2  & 78.17 & 62.38 & 68.95 & 77.76 & 75.29 \\
&   &                   &   & Qwen-3.8-Max & 86.87 & 72.68 & 80.06 & 82.30 & 80.70 \\
\midrule
\multirow{11}{*}{Qwen-3.6-27B} & \bvline{11}{3}
& No Train       & \grule{1} & no train & 74.73 & 54.91 & 64.73 & 71.59 & 71.26 \\
\cmidrule(lr){5-10}
&   & \multirow{4}{*}{Single} & \grule{4}
                            & glm\_base & 75.05 & 59.65 & 65.11 & 71.85 & 67.29 \\
&   &               &           & glm\_pro  & 75.30 & 56.05 & 65.28 & 73.38 & 73.76 \\
&   &               &           & glm\_max  & 74.64 & 62.16 & 66.83 & 73.17 & 73.72 \\
&   &               &           & qwen      & 80.86 & 68.20 & 70.75 & 77.88 & 76.23 \\
\cmidrule(lr){5-10}
&   & \multirow{3}{*}{Mixed} & \grule{3}
                            & qwen+glm\_pro          & 79.51 & 64.69 & 70.95 & 78.39 & 78.79 \\
&   &               &           & qwen+glm\_max          & 82.84 & 67.96 & 72.87 & 77.20 & 76.52 \\
&   &               &           & qwen+glm\_pro+glm\_max & 79.97 & 69.42 & 67.61 & 78.52 & 77.82 \\
\cmidrule(lr){5-10}
&   & \multirow{3}{*}{Mixed+Filter} & \grule{3}
                            & qwen+glm\_pro (filter) & 80.57 & 67.56 & 71.27 & 77.88 & 78.94 \\
&   &               &           & qwen+glm\_max (filter) & 81.73 & 68.31 & 72.84 & 77.59 & 76.49 \\
&   &               &           & qwen+glm\_pro+glm\_max (filter) & 80.99 & 66.96 & 69.89 & 79.50 & 79.52 \\
\midrule
\multirow{11}{*}{Qwen-3.6-35B-A3B} & \bvline{11}{3}
& No Train       & \grule{1} & no train & 67.72 & 53.50 & 58.00 & 66.58 & 68.72 \\
\cmidrule(lr){5-10}
&   & \multirow{4}{*}{Single} & \grule{4}
                            & glm\_base & 70.14 & 59.07 & 62.11 & 69.16 & 66.47 \\
&   &               &           & glm\_pro  & 72.38 & 61.99 & 60.44 & 69.38 & 71.43 \\
&   &               &           & glm\_max  & 73.43 & 60.65 & 62.87 & 66.83 & 70.63 \\
&   &               &           & qwen      & 74.11 & 57.91 & 63.40 & 70.28 & 68.85 \\
\cmidrule(lr){5-10}
&   & \multirow{3}{*}{Mixed} & \grule{3}
                            & qwen+glm\_pro          & 77.03 & 64.11 & 66.62 & 74.61 & 74.93 \\
&   &               &           & qwen+glm\_max          & 75.61 & 61.79 & 66.16 & 70.56 & 72.16 \\
&   &               &           & qwen+glm\_pro+glm\_max & 76.57 & 64.48 & 65.69 & 73.98 & 75.32 \\
\cmidrule(lr){5-10}
&   & \multirow{3}{*}{Mixed+Filter} & \grule{3}
                            & qwen+glm\_pro (filter) & 78.07 & 62.50 & 66.67 & 75.95 & 76.11 \\
&   &               &           & qwen+glm\_max (filter) & 74.05 & 61.41 & 63.30 & 71.00 & 73.24 \\
&   &               &           & qwen+glm\_pro+glm\_max (filter) & 76.91 & 61.44 & 66.83 & 74.05 & 74.80 \\
\bottomrule
\end{tabular}
\end{table*}

\begin{table*}[t]
\centering
\small
\setlength{\tabcolsep}{2.5pt}
\caption{Per-dimension PresentBench scores for two target models under different teacher references and supervised fine-tuning configurations. CQ: content quality; VD: visual design; Comp.: material completeness (coverage of source material); Corr.: material correctness (data/citation fidelity); Cons.: per-slide consistency (page-level alignment with source material).}
\label{tab:sft_per_dimension}
\begin{tabular}{lclclccccc}
\toprule
\multirow{2}{*}{\textbf{Target Model}} & & \multirow{2}{*}{\textbf{Group}} & & \multirow{2}{*}{\textbf{Setting}} & \multicolumn{5}{c}{\textbf{PresentBench Dimensions (\%)}} \\
\cmidrule(lr){6-10}
& & & & & \textbf{CQ} & \textbf{VD} & \textbf{Comp.} & \textbf{Corr.} & \textbf{Cons.} \\
\midrule
\multirow{2}{*}{Teacher} & \bvline{2}{0} & \multirow{2}{*}{--} & \grule{2}
                            & GLM-5.2 & 93.31 & 65.40 & 80.36 & 72.71 & 62.40 \\
&   &                   &   & Qwen-3.8-Max & 95.57 & 76.08 & 85.73 & 82.18 & 72.97 \\
\midrule
\multirow{11}{*}{Qwen-3.6-27B} & \bvline{11}{3}
& No Train       & \grule{1} & no train & 91.92 & 60.18 & 76.42 & 66.85 & 54.93 \\
\cmidrule(lr){5-10}
&   & \multirow{4}{*}{Single} & \grule{4}
                            & glm\_base & 91.08 & 61.76 & 76.97 & 66.70 & 53.87 \\
&   &               &           & glm\_pro  & 91.14 & 62.33 & 78.77 & 67.97 & 56.03 \\
&   &               &           & glm\_max  & 92.82 & 67.70 & 76.30 & 66.87 & 53.93 \\
&   &               &           & qwen      & 94.79 & 70.71 & 82.72 & 75.65 & 57.58 \\
\cmidrule(lr){5-10}
&   & \multirow{3}{*}{Mixed} & \grule{3}
                            & qwen+glm\_pro          & 94.50 & 70.09 & 82.80 & 73.72 & 60.01 \\
&   &               &           & qwen+glm\_max          & 95.89 & 71.97 & 82.23 & 76.12 & 60.56 \\
&   &               &           & qwen+glm\_pro+glm\_max & 95.95 & 69.67 & 82.59 & 73.09 & 58.77 \\
\cmidrule(lr){5-10}
&   & \multirow{3}{*}{Mixed+Filter} & \grule{3}
                            & qwen+glm\_pro (filter) & 96.79 & 71.35 & 82.03 & 74.33 & 59.16 \\
&   &               &           & qwen+glm\_max (filter) & 95.66 & 70.29 & 82.69 & 76.84 & 59.90 \\
&   &               &           & qwen+glm\_pro+glm\_max (filter) & 96.18 & 70.51 & 83.29 & 74.46 & 61.09 \\
\midrule
\multirow{11}{*}{Qwen-3.6-35B-A3B} & \bvline{11}{3}
& No Train       & \grule{1} & no train & 87.56 & 51.31 & 74.57 & 60.95 & 47.89 \\
\cmidrule(lr){5-10}
&   & \multirow{4}{*}{Single} & \grule{4}
                            & glm\_base & 88.41 & 58.18 & 77.32 & 64.28 & 45.57 \\
&   &               &           & glm\_pro  & 88.85 & 62.90 & 77.95 & 64.60 & 46.18 \\
&   &               &           & glm\_max  & 90.34 & 63.32 & 76.26 & 63.91 & 46.74 \\
&   &               &           & qwen      & 88.26 & 62.04 & 82.67 & 68.54 & 40.69 \\
\cmidrule(lr){5-10}
&   & \multirow{3}{*}{Mixed} & \grule{3}
                            & qwen+glm\_pro          & 92.88 & 66.48 & 82.16 & 70.97 & 51.47 \\
&   &               &           & qwen+glm\_max          & 90.81 & 66.78 & 80.96 & 69.34 & 43.50 \\
&   &               &           & qwen+glm\_pro+glm\_max & 93.63 & 67.62 & 81.24 & 67.35 & 52.33 \\
\cmidrule(lr){5-10}
&   & \multirow{3}{*}{Mixed+Filter} & \grule{3}
                            & qwen+glm\_pro (filter) & 94.37 & 68.68 & 82.74 & 69.70 & 52.65 \\
&   &               &           & qwen+glm\_max (filter) & 88.97 & 64.55 & 80.23 & 68.18 & 46.36 \\
&   &               &           & qwen+glm\_pro+glm\_max (filter) & 94.05 & 66.53 & 80.36 & 68.44 & 53.52 \\
\bottomrule
\end{tabular}
\end{table*}

% \subsection{Training Time}
% SFT training remains computationally tractable across varying data regimes and backbone architectures. With training sets ranging from roughly 3.7K to 7.1K trajectories, the dense backbone typically converges within about three to five days, while the MoE backbone requires around four to five days, indicating stable large-scale training efficiency.

\subsection{Supplementary Results for SFT}
Table~\ref{tab:sft_per_module} reports the performance of the teacher models and the student models before and after training on the five domain-specific subsets of PresentBench. Table~\ref{tab:sft_per_dimension} presents their score rates over the five evaluation dimensions of PresentBench. Table~\ref{tab:sft_score_matrix_all} summarizes their aesthetic metric scores on PresentBench and the PPT subset of GDPVal.

%% file: sections/07_04_appendix_sft_ae.tex
\begin{table*}[t]
\centering
\footnotesize
\setlength{\tabcolsep}{4pt}
\caption{Aspect-level score matrix on PresentBench and the PPT subset of GDPVal for supervised fine-tuning experiments, covering teacher models and two student model families (Qwen-3.6-27B and Qwen-3.6-35B-A3B). Teacher models are reported once at the top as references. Within each student family, rows correspond to the no-train baseline and SFT results under different training-data configurations. The best value in each column is boldfaced within each student family. Asp.\ = aspect-ratio compliance, Ws.\ = whitespace adequacy, Ele.\ = element-collision avoidance, Bal.\ = visual balance, Font = font readability, and Total = overall score.}

\label{tab:sft_score_matrix_all}
\resizebox{\textwidth}{!}{%
\begin{tabular}{llcccccccccccc}
\toprule
\multirow{2}{*}{\textbf{Group}} & \multirow{2}{*}{\textbf{Model}}
& \multicolumn{6}{c}{\textbf{PresentBench}} & \multicolumn{6}{c}{\textbf{GDPVal (PPT subset)}} \\
\cmidrule(lr){3-8} \cmidrule(lr){9-14}
& & \textbf{Asp.} & \textbf{Ws.} & \textbf{Ele.} & \textbf{Bal.} & \textbf{Font} & \textbf{Total}
  & \textbf{Asp.} & \textbf{Ws.} & \textbf{Ele.} & \textbf{Bal.} & \textbf{Font} & \textbf{Total} \\
\midrule
\multirow{2}{*}{Teacher}
& glm-5.2  & 0.99 & 0.87 & 0.79 & 0.54 & 0.95 & 4.13 & 0.94 & 0.84 & 0.62 & 0.50 & 0.77 & 3.67 \\
& Qwen-3.8-Max & 1.00 & 0.95 & 0.67 & 0.53 & 0.93 & 4.08 & 1.00 & 0.90 & 0.50 & 0.56 & 0.63 & 3.59 \\
\midrule
\multirow{2}{*}{Student}
& Qwen-3.6-27B (no train)      & 0.99 & 0.76 & 0.65 & 0.48 & 0.94 & 3.81 & 0.93 & 0.69 & 0.54 & 0.43 & 0.66 & 3.26 \\
& Qwen-3.6-35B-A3B (no train)  & 0.98 & 0.80 & 0.60 & 0.50 & 0.97 & 3.85 & 0.95 & 0.78 & 0.46 & 0.51 & 0.64 & 3.34 \\
\midrule
\multirow{10}{*}{Qwen-3.6-27B (SFT)}
& glm-base                           & 0.99 & 0.83 & 0.76 & 0.48 & 0.87 & 3.93 & \textbf{0.95} & 0.79 & 0.65 & 0.55 & 0.75 & \textbf{3.70} \\
& glm-pro                            & \textbf{1.00} & 0.86 & 0.76 & 0.56 & 0.96 & 4.12 & 0.91 & 0.87 & 0.57 & 0.49 & 0.82 & 3.67 \\
& glm-max                            & \textbf{1.00} & 0.86 & 0.62 & 0.54 & \textbf{0.97} & 3.99 & 0.66 & 0.85 & \textbf{0.70} & 0.47 & \textbf{0.96} & 3.64 \\
& qwen                           & 0.99 & \textbf{0.97} & 0.51 & 0.58 & 0.95 & 4.00 & 0.73 & 0.92 & 0.40 & 0.47 & 0.64 & 3.16 \\
& qwen+glm-pro                       & 0.99 & 0.89 & 0.70 & 0.56 & 0.94 & 4.09 & 0.78 & \textbf{0.95} & 0.33 & 0.45 & 0.72 & 3.22 \\
& qwen+glm-max                       & 0.99 & \textbf{0.97} & 0.57 & 0.58 & 0.95 & 4.06 & 0.70 & 0.89 & 0.51 & 0.38 & 0.65 & 3.13 \\
& qwen+glm-pro+glm-max              & 0.99 & 0.88 & 0.72 & 0.57 & 0.95 & 4.13 & 0.86 & 0.88 & 0.49 & \textbf{0.57} & 0.79 & 3.57 \\
& qwen+glm-pro (filtered)            & 0.99 & 0.89 & \textbf{0.78} & 0.57 & \textbf{0.97} & \textbf{4.20} & 0.74 & 0.87 & 0.58 & 0.49 & 0.71 & 3.38 \\
& qwen+glm-max (filtered)            & 0.99 & \textbf{0.97} & 0.58 & \textbf{0.59} & 0.94 & 4.06 & 0.62 & 0.91 & 0.39 & 0.49 & 0.67 & 3.08 \\
& qwen+glm-pro+glm-max (filtered)   & \textbf{1.00} & 0.88 & 0.74 & 0.57 & 0.95 & 4.14 & 0.70 & 0.94 & 0.57 & 0.44 & 0.70 & 3.35 \\
\midrule
\multirow{10}{*}{Qwen-3.6-35B-A3B (SFT)}
& glm-base                           & 0.99 & 0.83 & 0.72 & 0.51 & 0.90 & 3.93 & 0.87 & 0.85 & 0.52 & \textbf{0.57} & 0.78 & 3.59 \\
& glm-pro                            & 0.99 & 0.86 & \textbf{0.75} & \textbf{0.58} & 0.95 & 4.13 & 0.80 & 0.68 & 0.37 & 0.55 & 0.81 & 3.22 \\
& glm-max                            & \textbf{1.00} & 0.87 & 0.60 & 0.53 & \textbf{0.97} & 3.97 & 0.91 & 0.84 & \textbf{0.58} & 0.53 & \textbf{0.95} & \textbf{3.80} \\
& qwen                           & 0.97 & \textbf{0.96} & 0.47 & \textbf{0.58} & 0.95 & 3.94 & 0.77 & 0.75 & 0.24 & 0.54 & 0.70 & 3.01 \\
& qwen+glm-pro                       & 0.99 & 0.89 & 0.66 & 0.56 & 0.95 & 4.06 & 0.72 & 0.83 & 0.57 & 0.51 & 0.75 & 3.38 \\
& qwen+glm-max                       & 0.99 & \textbf{0.96} & 0.53 & 0.57 & 0.95 & 4.00 & 0.74 & \textbf{0.88} & 0.48 & 0.48 & 0.60 & 3.17 \\
& qwen+glm-pro+glm-max              & \textbf{1.00} & 0.86 & 0.74 & 0.57 & 0.95 & 4.11 & 0.75 & 0.82 & 0.53 & 0.52 & 0.81 & 3.43 \\
& qwen+glm-pro (filtered)            & 0.99 & 0.89 & 0.74 & \textbf{0.58} & 0.95 & \textbf{4.16} & 0.88 & 0.73 & 0.49 & 0.54 & 0.80 & 3.44 \\
& qwen+glm-max (filtered)            & 0.99 & 0.95 & 0.51 & 0.57 & 0.95 & 3.98 & 0.83 & 0.83 & 0.49 & 0.56 & 0.77 & 3.48 \\
& qwen+glm-pro+glm-max (filtered)   & 0.99 & 0.86 & 0.74 & 0.57 & \textbf{0.97} & 4.13 & 0.93 & 0.77 & 0.52 & \textbf{0.57} & 0.80 & 3.59 \\
\bottomrule
\end{tabular}%
}
\end{table*}

%% file: sections/07_05_appendix_rl_reward.tex
\section{Reward Design Details}
\label{sec:reward}

While SFT is effective at aligning generated presentations with teacher trajectories, its supervision is fundamentally imitation-based and therefore insufficient for learning reliable visual-quality signals. In particular, although SFT equips the model with the basic procedural competence required for slide construction, it does not explicitly optimize the visual criteria that largely determine users' first impressions of a presentation~\cite{pan2026aeslides}. In our observations, post-SFT presentations still exhibit defects such as content overflow, excessive whitespace, visual imbalance, and element overlap. These flaws substantially reduce readability and perceived quality. We therefore introduce reinforcement learning with \emph{verifiable} visual rewards to refine presentation aesthetics beyond text-only supervision.

Rather than relying on a learned critic or VLM judge, we decompose aesthetic quality into five deterministic and interpretable subproblems, each implemented as a rule-based function over rendered slides. Let $x$ denote a rendered slide and let $R_m(x)\in[0,1]$ denote the reward for metric $m$. The final slide-level reward is the unweighted mean
\begin{equation}
R_{\mathrm{slide}}(x)=\frac{1}{5}\sum_{m\in\{\mathrm{asp},\mathrm{ws},\mathrm{ov},\mathrm{cen},\mathrm{font}\}} R_m(x),
\end{equation}
and the deck-level reward is obtained by averaging over slides. This design yields a cheap, stable, and fully auditable optimization target.

\subsection{Render Infrastructure}

All metrics are computed on fully rendered slides rather than raw HTML, so that both DOM structure and final visual evidence are available after CSS layout, responsive sizing, and font resolution have been faithfully applied. To support this process, we build a distributed rendering service based on headless Chromium. Depending on the metric, the renderer returns either (i) an adaptive-height rendering result for aspect-ratio analysis, (ii) a Playwright screenshot for image-based whitespace estimation, or (iii) a rich semantic bounding-box tree (\texttt{bbox\_tree}) containing node types, element boxes, text-line rectangles from \texttt{Range.getClientRects()}, and resolved style attributes such as font size and background color. This decoupled design enables accurate reward computation on final visual outputs while maintaining high rollout throughput during RL.

\subsection{Verifiable Aesthetic Metrics}
\label{subsec:metrics}

We target five recurring visual failure modes in automatic slide generation: aspect-ratio distortion, excessive whitespace, element collision, visual imbalance, and poor font readability. Each metric is normalized to $[0,1]$, where larger is better.

\paragraph{(1) Aspect-ratio compliance.}
This metric evaluates whether a slide preserves the canonical $16{:}9$ presentation ratio after responsive rendering. Given HTML content, we render the slide at a fixed viewport width $W=1280$ with adaptive height $H$, and compute the realized aspect ratio
\begin{equation}
\mathrm{AR}=\frac{W}{H}.
\end{equation}
Let $\tau=1280/720$ denote the target ratio. We use a non-symmetric exponential penalty that tolerates mild deviations but penalizes vertically elongated layouts more heavily, since such cases usually indicate overflow. Concretely, with
\begin{equation}
e=\log\!\left(\frac{\mathrm{AR}}{\tau}\right), \qquad
\delta_{\mathrm{tall}}=\max(-e-m,0),
\end{equation}
where $m$ is a small margin, the reward is
\begin{equation}
R_{\mathrm{asp}}
=\exp\!\left(-\alpha e^2-\beta \delta_{\mathrm{tall}}^2\right).
\end{equation}
In our implementation, $\alpha=16$, $\beta=64$, and $m=0.04$. Thus, the reward equals $1$ at the target ratio, decays smoothly under small mismatch, and drops more sharply when the slide becomes too tall.

\paragraph{(2) Whitespace adequacy.}
This metric measures whether a slide contains sufficient visual content rather than large empty regions. We first render the HTML slide using the same Playwright-based screenshot pipeline employed in our HTML-to-PDF conversion. The screenshot is converted to grayscale, smoothed by a Gaussian kernel, and processed with a large box filter to estimate local variance:
\begin{equation}
\sigma_{\mathrm{loc}}(p)=\sqrt{\mathbb{E}_{\mathcal{N}(p)}[I^2]-\mathbb{E}_{\mathcal{N}(p)}[I]^2},
\end{equation}
where $I$ is the grayscale image and $\mathcal{N}(p)$ is a local neighborhood around pixel $p$. After clipping and normalization, we binarize the map to obtain a foreground indicator
\begin{equation}
M(p)=\mathbb{I}\!\left(\frac{\min(\sigma_{\mathrm{loc}}(p),c)}{c}\ge t\right),
\end{equation}
where $c$ is the standard-deviation clip value and $t$ is a binary threshold. We then crop a central region of interest to suppress margins and compute the effective content ratio
\begin{equation}
r_{\mathrm{content}}=\frac{1}{|\Omega_{\mathrm{crop}}|}\sum_{p\in\Omega_{\mathrm{crop}}} M(p).
\end{equation}
The reward is obtained via a smoothstep mapping
\begin{equation}
R_{\mathrm{ws}}=
\mathrm{SmoothStep}(r_{\mathrm{content}}; l,u)=
\begin{cases}
0, & r_{\mathrm{content}}\le l,\\
3z^2-2z^3, & l<r_{\mathrm{content}}<u,\ z=\frac{r_{\mathrm{content}}-l}{u-l},\\
1, & r_{\mathrm{content}}\ge u,
\end{cases}
\end{equation}
with $l=0.2$ and $u=0.7$. Intuitively, nearly blank slides receive low reward, while content-rich slides receive high reward.

\paragraph{(3) Element-collision avoidance.}
This metric penalizes undesirable overlap among semantically meaningful elements as well as boundary violations. Given a semantic \texttt{bbox\_tree}, we collect two classes of rectangles: text-line rectangles from \texttt{Range.getClientRects()} and leaf-element boxes for visual objects such as figures, icons, charts, and images. For a pair of rectangles $a$ and $b$ with areas $|a|$ and $|b|$, let
\begin{equation}
A_{\cap}(a,b)=|a\cap b|,\qquad
\rho(a,b)=\frac{A_{\cap}(a,b)}{\min(|a|,|b|)}.
\end{equation}
A pair is counted as a collision if it has no ancestor--descendant relation in the DOM tree and satisfies
\begin{equation}
A_{\cap}(a,b)\ge \eta_A
\quad \text{and} \quad
\rho(a,b)\ge \eta_\rho,
\end{equation}
where $\eta_A=10$ px$^2$ and $\eta_\rho=0.10$. In addition, any rectangle whose distance to the viewport boundary is smaller than a margin $\mu=8$ px is counted as a boundary violation. Let $N_{\mathrm{ov}}$ denote the number of valid overlap collisions and let $N_{\mathrm{bd}}$ denote the number of boundary violations. The total error count is
\begin{equation}
N_{\mathrm{err}}=N_{\mathrm{ov}}+N_{\mathrm{bd}},
\end{equation}
and the reward is
\begin{equation}
R_{\mathrm{ov}}=\max\!\left(0,\ 1-\lambda N_{\mathrm{err}}\right),
\end{equation}
with $\lambda=0.5$. Hence, the reward is $1$ when no collision is detected, decreases linearly with errors, and becomes $0$ after sufficiently many violations.

\paragraph{(4) Visual balance.}
This metric evaluates whether visual mass is distributed harmoniously across the slide while remaining tolerant to common asymmetric designs such as left-image-right-text layouts. Using the semantic \texttt{bbox\_tree}, we first remove title-like blocks near the top and footer-like blocks near the bottom, then retain content elements with area above a minimal threshold. Let each retained block $i$ have area weight $w_i$ and center $(x_i,y_i)$. We partition the slide into left/right and top/bottom halves and compute
\begin{equation}
W_L=\sum_{i:x_i<1/2} w_i,\quad
W_R=\sum_{i:x_i\ge 1/2} w_i,\quad
W_T=\sum_{i:y_i<1/2} w_i,\quad
W_B=\sum_{i:y_i\ge 1/2} w_i.
\end{equation}
The raw horizontal and vertical balance scores are
\begin{equation}
b_{LR}^{\mathrm{raw}}=1-\frac{|W_L-W_R|}{W_L+W_R},\qquad
b_{TB}^{\mathrm{raw}}=1-\frac{|W_T-W_B|}{W_T+W_B}.
\end{equation}
To make the metric mildly tolerant rather than overly strict, we use a power penalty with exponent $p=1.5$:
\begin{equation}
b_{LR}=(b_{LR}^{\mathrm{raw}})^p,\qquad
b_{TB}=(b_{TB}^{\mathrm{raw}})^p.
\end{equation}
We further introduce center-alignment factors $c_{LR},c_{TB}\in[0,1]$, which measure whether the weighted content centers within each partition remain reasonably close to the central axes. The horizontal and vertical scores are
\begin{equation}
s_{LR}=b_{LR}\big((1-\gamma)+\gamma c_{LR}\big),\qquad
s_{TB}=b_{TB}\big((1-\gamma)+\gamma c_{TB}\big),
\end{equation}
where $\gamma=0.3$. The combined balance score is
\begin{equation}
b= \omega_{LR}s_{LR}+\omega_{TB}s_{TB},
\end{equation}
with $\omega_{LR}=0.6$ and $\omega_{TB}=0.4$. Finally, we map it through a smoothstep function,
\begin{equation}
R_{\mathrm{cen}}=\mathrm{SmoothStep}(b;0,0.85),
\end{equation}
which yields a tolerant yet discriminative reward for slide-level composition quality.

\paragraph{(5) Font readability.}
This metric evaluates whether textual content remains legible under realistic presentation conditions. From the \texttt{bbox\_tree}, we collect all text-like nodes with font size $f_i$, area $a_i$, and word count $n_i$. Text nodes are categorized by size: unreadable ($f_i<12$ px), small ($12\le f_i<14$ px), ideal ($14\le f_i\le 52$ px), mildly large ($52<f_i\le 72$ px), and excessively large ($f_i>72$ px). Very small footnote-like regions are exempted, and for the title slide we disable large-font penalties. Let $A_{\mathrm{ideal}}$, $A_{\mathrm{small}}$, and $A_{\mathrm{unread}}$ denote the total area of ideal, small, and unreadable text, respectively, and let
\begin{equation}
A_{\mathrm{tot}}=\sum_i a_i.
\end{equation}
The base readable ratio is
\begin{equation}
r_{\mathrm{read}}=\frac{A_{\mathrm{ideal}}+0.5A_{\mathrm{small}}}{A_{\mathrm{tot}}},
\end{equation}
and the base reward is
\begin{equation}
\widetilde{R}_{\mathrm{font}}=
\mathrm{clip}\!\left(
r_{\mathrm{read}}\Big(1-2\frac{A_{\mathrm{unread}}}{A_{\mathrm{tot}}}\Big),\ 0,\ 1
\right).
\end{equation}
We then apply two hard penalties: one for the existence of any non-footnote sub-12px text, and another for dense small-font blocks (defined as text blocks with font size at most 12px and at least 35 consecutive words). Let $\mathbb{I}_{<12}$ indicate whether any such sub-12px text exists, and let $N_{\mathrm{dense}}$ denote the number of dense small-text blocks. The final reward is
\begin{equation}
R_{\mathrm{font}}=
\mathrm{clip}\!\left(
\widetilde{R}_{\mathrm{font}}
-0.5\,\mathbb{I}_{<12}
-0.5\,N_{\mathrm{dense}},\ 0,\ 1
\right).
\end{equation}
If no text is present, we assign a neutral score of $0.5$.

\paragraph{Reward aggregation.}
Since the five metrics are complementary and normalized to a common scale, we aggregate them using an unweighted mean to obtain both slide-level and deck-level rewards. This simple design avoids manual weighting and provides a transparent, stable, and verifiable optimization objective for reinforcement learning:
\begin{equation}
R_{\mathrm{deck}}=\frac{1}{T}\sum_{t=1}^{T}R_{\mathrm{slide}}(x_t),
\end{equation}
where $x_t$ is the rendered output of slide $t$ in a deck of length $T$.

%% file: sections/07_06_appendix_rl_training.tex
\section{RL Implementation Details}

\subsection{Experimental Setup}

We provide a comprehensive account of the experimental setup in Tables~\ref{tab:appendix_train_opt}, \ref{tab:appendix_parallel_eval}, and \ref{tab:appendix_render_metrics}. The two backbones share the same reward formulation, optimization objective, and stabilization strategy, and differ primarily in parallelization layout and scale. In particular, both backbones use the same importance-ratio parameterization: a GSPO-style segment-level ratio, i.e.\ a per-segment geometric mean of the token log-ratio, where segments are the contiguous policy-generated spans of a multi-turn trajectory. Relative to a token-level ratio, this prevents any single outlier token from dominating the update.

Both policies are trained with GRPO using a group-mean baseline and batch-level standard-deviation normalization, following a reward-decoupled formulation. We employ a token-mean policy loss, asymmetric PPO clipping, and an explicit KL anchor to the SFT initialization. The anchor itself is studied as a controlled pair of runs, with and without the KL-loss term, while all remaining settings are held fixed within each backbone. To improve robustness, especially during the early phase of training, we additionally recompute behavior log-probabilities with the training engine and apply token-level truncated importance sampling to mitigate train--inference mismatch, discarding tokens whose train--inference ratio falls outside a prescribed range. These mechanisms are critical for maintaining stable optimization in long-horizon agentic rollouts with structured rendering rewards.

\begin{table}[t]
\centering
\small
\setlength{\tabcolsep}{8pt}
\caption{Training, rollout, and optimization configurations. All listed hyperparameters are shared by the two backbones (Dense-27B and MoE-35B-A3B); backbone-specific settings are reported in Table~\ref{tab:appendix_parallel_eval}. The KL anchor is ablated as a matched pair (anchored vs.\ non-anchored); the coefficient applies to the anchored arm.}
\label{tab:appendix_train_opt}
\begin{tabularx}{0.85\linewidth}{
>{\raggedright\arraybackslash}X
>{\raggedright\arraybackslash}X
}
\toprule
\textbf{Hyperparameter} & \textbf{Shared Value} \\
\midrule
Advantage estimator & GRPO \\
% Optimization steps (max) & 256 \\
Rollout batch size (rollouts / step) & 64 \\
Environment groups / step & 8 \\
Samples per prompt (group size) & 8 \\
Rollout temperature & 1.0 \\
Top-p / Top-k & 1.0 / 50 \\
Total sequence length & 131,072 \\
Context window & 122,880 \\
Per-turn response cap (max new tokens) & 8,192 \\
Max actions per trajectory & 128 \\
Learning rate & 3e-6 \\
LR schedule / warmup & linear / 15 steps \\
Weight decay & 0.0 \\
epochs & 1 \\
Loss aggregation & token-mean \\
Advantage clip & 20 \\
Max gradient norm & 1.0 \\
Clip range $(\epsilon_{\mathrm{low}}, \epsilon_{\mathrm{high}})$ & (0.1, 0.28) \\
Dual-clip & enabled \\
Entropy loss coefficient & 0.0 \\
Whiten advantages & false \\
Reward-side KL coefficient & 0.0 \\
KL-loss anchor & on (k3) vs.\ off (ablated pair) \\
KL-loss coefficient $\beta$ & 1e-3 (anchored arm) \\
Reward normalization & group mean / batch std \\
Old-logprob recomputation & enabled \\
TIS importance weight & token, $\leq 1.2$ \\
TIS mismatch filter & drop if ratio $\notin [0.5, 2.0]$ \\
Importance ratio type & segment (GSPO-style) \\
\bottomrule
\end{tabularx}
\end{table}

\begin{table}[t]
\centering
\small
\caption{Parallelism and hardware configurations for the two backbones.}
\label{tab:appendix_parallel_eval}
\setlength{\tabcolsep}{3pt}
\renewcommand{\arraystretch}{1.0}
\begin{tabularx}{0.9\linewidth}{
>{\raggedright\arraybackslash}X
>{\centering\arraybackslash}X
>{\centering\arraybackslash}X
}
\toprule
\textbf{Hyperparameter} & \textbf{Dense-27B} & \textbf{MoE-35B-A3B} \\
\midrule
Tensor parallel & 8 & 1 \\
Pipeline parallel & 1 & 4 \\
Context parallel & 2 & 8 \\
Expert parallel & 1 & 8 \\
Data parallel & 2 & 4 \\
Gradient accumulation steps & 32 & 16 \\
Per-device train batch & 1 & 1 \\
Recompute & full, uniform, 1 layer & full, uniform, 1 layer \\
Cross-entropy loss fusion & true & true \\
Dynamic batching & on & on \\
vLLM rollout TP & 4 & 4 \\
Precision & bf16 & bf16 \\
\bottomrule
\end{tabularx}
\end{table}

\begin{table}[t]
\centering
\small
\caption{Render infrastructure and verifiable aesthetic metrics used in reinforcement learning.}
\label{tab:appendix_render_metrics}
\begin{minipage}{0.85\textwidth}
\centering
\setlength{\tabcolsep}{14pt}
\renewcommand{\arraystretch}{1.0}
\begin{tabularx}{\linewidth}{
>{\raggedright\arraybackslash}p{5.0cm}
>{\raggedright\arraybackslash}X
}
\toprule
\textbf{Component / Parameter} & \textbf{Value} \\
\midrule
Renderer & headless Chromium (Playwright) \\
Adaptive-height render width & 1280 px \\
Fixed render viewport & 1280 $\times$ 720, device scale factor 2 \\
Aspect-ratio target $\tau$ & 16:9 (1280/720) \\
Aspect-ratio $\alpha$ / $\beta$ / margin $m$ & 16.0 / 64.0 / 0.04 \\
Whitespace Gaussian kernel & 21 $\times$ 21 \\
Whitespace box kernel (H $\times$ V) & 201 $\times$ 151 \\
Whitespace std clip / binary threshold & 50.0 / 0.15 \\
Whitespace crop ratios (T/B/L/R) & 0.10 / 0.08 / 0.08 / 0.08 \\
Whitespace smoothstep (lower, upper) & (0.2, 0.7) \\
Collision boundary margin & 8 px \\
Collision min text-rect / element area & 20 px$^2$ / 100 px$^2$ \\
Collision min overlap area / ratio & 10 px$^2$ / 0.10 \\
Collision per-error penalty & 0.5 \\
Visual-balance title / footer threshold & 0.18 / 0.92 \\
Visual-balance power / horizontal & 1.5 / 0.6 \\
Vertical weight & 0.4 \\
Visual center penalty weight & 0.3 \\
Visual smoothstep (lower, upper) & (0.0, 0.85) \\
Font readable / ideal / maximum px & 12 / 14--52 / 72 \\
Font footnote max area ratio & 0.01 \\
Font sub-12px / dense-block penalty & 0.5 each \\
Reward aggregation & unweighted mean of 5 metrics in $[0,1]$ \\
\bottomrule
\end{tabularx}
\end{minipage}
\end{table}

\subsection{Training Strategy Design}
Each optimization step collects $G_{\mathrm{grp}}\!=\!8$ prompts with group size
$8$ (a total of $64$ rollouts), where the in-group samples share the prompt and
seed to support the group-relative baseline. Rollouts are sampled at temperature
$1.0$ (top-$k\!=\!50$), with a total sequence budget of $131{,}072$ tokens
($122{,}880$ context $+$ an $8{,}192$-token per-turn response cap) and up to
$128$ agent actions per trajectory. We train with a linear-warmup learning rate
of $3\times10^{-6}$ and gradient clipping at $1.0$. We validate the recipe on
two backbones under identical rewards and RL objectives: a $27$B dense
model (tensor-parallel $8$ $\times$ context-parallel $2$, with the rollout
engine colocated on the same GPU set) and a $35$B-A3B MoE model
(tensor-parallel $1$ $\times$ pipeline-parallel $4$ $\times$ context-parallel
$8$ $\times$ expert-parallel $8$, colocated). For each backbone we
further run a matched pair of experiments that differ only in whether the KL
anchor to the SFT initialization is active, isolating its contribution to
training stability.

% \subsection{Training Time}
% The RL stage also remains practically tractable despite the long-horizon nature of slide generation. On average, 100 training steps require about five to six days for the dense backbone and around four to five days for the MoE backbone, reflecting stable large-scale optimization efficiency under tool-augmented rollout generation.

\subsection{Supplementary Results for RL}
Table~\ref{tab:sft_per_module} reports post-RL performance on the five domain-specific subsets of PresentBench, and Table~\ref{tab:sft_per_dimension} presents post-RL score rates across its five evaluation dimensions.

% Table~\ref{tab:rl_results_presentbench_gdpval} summarizes the overall RL results on PresentBench and the PPT subset of GDPVal.

To verify that our RL strategy not only optimizes the training objective but also aligns generated presentations more closely with human aesthetic preferences, we conduct a human evaluation with three annotators who are not involved in this project. Specifically, for each backbone, we compare the post-RL configuration achieving the highest aggregate score over the five aesthetic metrics against its pre-RL counterpart. We randomly sample 150 cases per backbone and ask the annotators, under a blind protocol, to choose the presentation with better visual quality and overall quality. The results show that, for the dense backbone, the post-RL model achieves win rates of 77.1\% in visual quality and 66.9\% in overall quality. For the MoE backbone, the corresponding win rates are 82.4\% and 71.8\%, respectively. These human evaluation results further validate the effectiveness of our training strategy.

%% file: sections/07_08_appendix_future_work.tex
\section{Future Work}

\paragraph{Reward modeling for content quality.}
A particularly important step toward this goal is the construction of reward models for content-related supervision. In the current framework, content quality is largely inherited from SFT initialization and is not yet directly optimized during RL. Future work should develop content reward models that assess dimensions such as factual faithfulness, material completeness, and slide-level semantic alignment. One practical route is to collect manually scored cases and train content reward models via supervised fine-tuning on these annotated examples, thereby distilling human judgments of presentation content quality into a learnable evaluation function. Such reward models could provide a scalable supervisory signal for content-aware RL, especially in aspects that are difficult to capture with purely rule-based criteria.

\paragraph{Algorithmic optimization.}
While our GRPO-based reinforcement learning framework yields consistent gains in aesthetic quality, its trajectory-level reward assignment remains inherently coarse for long-horizon presentation generation. A promising direction is to develop more expressive optimization paradigms, including step-level or hierarchical credit assignment~\cite{feng2026group,he2026hierarchy} and reinforcement learning integrated with on-policy distillation~\cite{liao2026multi,li2026rethinking}. By providing denser intermediate supervision and more context-aware attribution of final visual outcomes to individual actions, such methods may substantially improve optimization efficiency, stability, and scalability, ultimately enabling higher-quality and more reliable model optimization.

\paragraph{Personalized Presentation Generation}

While our framework optimizes for broadly preferred presentation quality, real-world presentation design is inherently heterogeneous across users, domains, and usage scenarios. A promising future direction is therefore personalized presentation generation, where supervision and optimization are conditioned on user-specific or scenario-specific preferences. This requires both preference-aware data construction and controllable post-training mechanisms, such as preference-conditioned fine-tuning and reward-conditioned reinforcement learning. Advancing in this direction would move presentation generation from general quality optimization toward truly adaptive presentation intelligence.